\documentclass{article}

\PassOptionsToPackage{numbers, compress}{natbib}

\usepackage[final]{neurips_2021}

\usepackage[utf8]{inputenc} 
\usepackage[T1]{fontenc}    
\usepackage{hyperref}       
\usepackage{url}            
\usepackage{booktabs}       
\usepackage{amsfonts}       
\usepackage{nicefrac}       
\usepackage{microtype}      
\usepackage{xcolor}         

\usepackage{microtype}
\usepackage{graphicx}
\usepackage{subcaption}
\usepackage{wrapfig}

\usepackage[algo2e]{algorithm2e}
\usepackage{amsmath}
\usepackage{amssymb}
\usepackage{multirow}
\usepackage{multicol}
\usepackage{graphicx}
\usepackage{float}
\usepackage{footmisc}
\usepackage{booktabs}
\usepackage{wrapfig}
\usepackage{xcolor}
\usepackage{soul}
\usepackage{dblfloatfix}
\usepackage[export]{adjustbox}
\DeclareMathOperator*{\argmax}{arg\,max}
\DeclareMathOperator*{\argmin}{arg\,min}

\makeatletter
\newcommand*{\addFileDependency}[1]{
  \typeout{(#1)}
  \@addtofilelist{#1}
  \IfFileExists{#1}{}{\typeout{No file #1.}}
}
\makeatother

\title{Stronger NAS with Weaker Predictors}

\author{%
  Junru Wu\textsuperscript{1}, Xiyang Dai\textsuperscript{2}, Dongdong Chen\textsuperscript{2}, Yinpeng Chen\textsuperscript{2}, \textbf{Mengchen Liu\textsuperscript{2}}, Ye Yu\textsuperscript{2},\\ \textbf{Zhangyang Wang\textsuperscript{3}}, \textbf{Zicheng Liu\textsuperscript{2}}, \textbf{Mei Chen\textsuperscript{2}}, \textbf{Lu Yuan\textsuperscript{2}}\\
  {\textsuperscript{1} Texas A\&M University, \textsuperscript{2}Microsoft Corporation, \textsuperscript{3}University of Texas at Austin} \\
  \small{\texttt{sandboxmaster@tamu.edu,
  \{xidai,dochen,yilche,mengcliu\}@microsoft.com,}}\\
  \small{\texttt{atlaswang@utexas.edu, \{yeyu1,zliu,mei.chen,luyuan\}@microsoft.com}}\\
}

\begin{document}

\maketitle

\begin{abstract}
\vspace{-0.5em}

Neural Architecture Search (NAS) often trains and evaluates a large number of architectures. Recent predictor-based NAS approaches attempt to alleviate such heavy computation costs with two key steps: sampling some architecture-performance pairs and fitting a proxy accuracy predictor. Given limited samples, these predictors, however, are far from accurate to locate top architectures due to the difficulty of fitting the huge search space. This paper reflects on a simple yet crucial question: \textit{if our final goal is to find the best architecture, do we really need to model the whole space well?}. We propose a paradigm shift from fitting the whole architecture space using one strong predictor, to progressively fitting a search path towards the high-performance sub-space through a set of weaker predictors. As a key property of the weak predictors, their probabilities of sampling better architectures keep increasing. Hence we only sample a few well-performed architectures guided by the previously learned predictor and estimate a new better weak predictor. This embarrassingly easy framework, dubbed \textbf{WeakNAS}, produces coarse-to-fine iteration to gradually refine the ranking of sampling space. Extensive experiments demonstrate that WeakNAS costs fewer samples to find top-performance architectures on NAS-Bench-101 and NAS-Bench-201. Compared to state-of-the-art (SOTA) predictor-based NAS methods, WeakNAS outperforms all with notable margins, e.g., requiring \textbf{at least 7.5x} less samples to find global optimal on NAS-Bench-101. WeakNAS can also absorb their ideas to boost performance more. Further, WeakNAS strikes the new SOTA result of 81.3\% in the ImageNet MobileNet Search Space. The code is available at: \url{https://github.com/VITA-Group/WeakNAS}.

\end{abstract}

\vspace{-1.5em}

\section{Introduction}
\vspace{-0.5em}

\begin{wrapfigure}{r}{0.37\textwidth}
\vspace{-0.5em}
\centering
\vspace{-3.5em}
\includegraphics[width=0.37\textwidth]{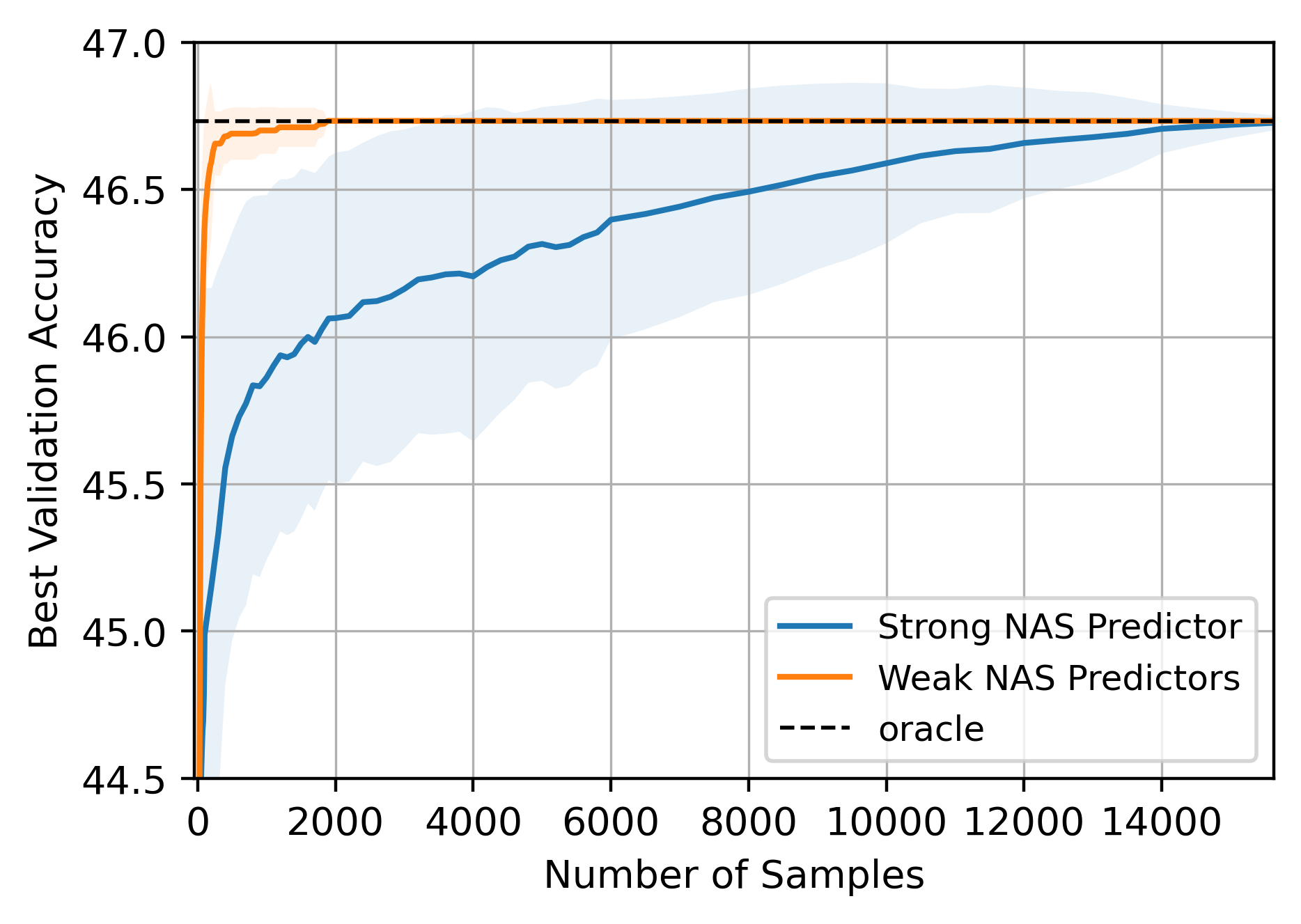}
\vspace{-2em}
\caption{\small{Comparison between our method using a set of weak predictors (iterative sampling), and a single strong predictor (random sampling) on NAS-Bench-201. For fair comparison, the NAS predictor in both methods adtops the same type of MLP described in \ref{sec:generalize}. Solid lines and shadows denote the mean and standard deviation (std), respectively.}}
\label{fig:teaser}
\vspace{-1.5em}
\end{wrapfigure}

Neural Architecture Search (NAS)  \citep{liu2018darts,luo2018neural,wu2019fbnet,howard2019searching,ning2020generic,wei2020npenas,wen2019neural,chau2020brp,luo2020neural,wang2020dc,dai2020data,yang2020hournas} methods aim to find the best network architecture by exploring the architecture-to-performance manifold, using reinforced-learning-based \citep{zoph2016neural}, evolution-based \citep{real2019regularized,yang2020cars} or gradient-based \citep{liu2018darts,hong2020dropnas} approaches. In order to cover the entire search space, they often train and evaluate a large number of architectures, leading to tremendous computation cost.

Recently, predictor-based NAS methods alleviate this problem with two key steps: one sampling step to sample some architecture-performance pairs, and another performance modeling step to fit the performance distribution by training a proxy accuracy predictor.  An in-depth analysis of existing methods \citep{luo2018neural} found that most of those methods \citep{ning2020generic,wei2020npenas,xu2019renas,wen2019neural,chau2020brp,luo2020neural,li2020neural} consider these two steps independently and attempt to model the performance distribution over the whole architecture space using a \textbf{\textit{strong}}\footnote{\label{strongweak}“Strong” vs “Weak” predictors: we name a “weak” predictor if it only predicts a local subspace of the search space thus can be associated with our iterative sampling scheme; such predictors therefore usually do not demand very heavily parameterized models. On the contrary, “strong” predictors predict the global search space and are often associated with uniform sampling. The terminology of strong versus weak predictors does not represent their number of parameters or the type of NAS predictor used. An overparameterized NAS predictor with our iterative sampling scheme may still be considered as a “weak” predictor.} predictor. However, since the architecture space is often exponentially large and highly non-convex, even a very strong predictor model has difficulty fitting the whole space given limited samples. Meanwhile, different types of predictors often demand handcraft design of the architecture representations to improve their performance.

This paper reflects on a fundamental question for predictor-based NAS: \textit{``if our final goal is to find the best architecture, do we really need to model the whole space well?''}. We investigate the alternative of utilizing a few \textbf{\textit{weak}}\footref{strongweak} predictors to fit small local spaces, and to progressively move the search space towards the subspace where good architecture resides. Intuitively, we assume the whole space could be divided into different sub-spaces, some of which are relatively good while some are relatively bad. We tend to choose the good ones while discarding the bad ones, which makes sure more samples will be focused on modeling only the good subspaces and then find the best architecture. It greatly simplifies the learning task of each predictor. Eventually, a line of progressively evolving weak predictors can connect a path to the best architecture.

We present a novel, general framework that requires only to estimate a series of weak predictors progressively along the search path, we denoted it as \textbf{WeakNAS} in the rest of the paper. To ensure moving towards the best architecture along the path, at each iteration, the sampling probability of better architectures keep increasing through the guidance of the previous weak predictor. Then, the consecutive weak predictors with better samples will be trained in the next iteration. We iterate until we arrive at an embedding subspace where the best architectures reside and can be accurately assessed by the final weak predictor.

Compared to the existing predictor-based NAS, our proposal represents a new line of attack and has several merits. First, since only weak predictors are required, it yields better sample efficiency. As shown in Figure \ref{fig:teaser}, it costs significantly fewer samples to find the top-performance architecture than using one strong predictor, and yields much lower variance in performance over multiple runs.
Second, it is flexible to the choices of architecture representation (e.g., different architecture embeddings) and predictor formulation (e.g., multilayer perceptron (MLP), gradient boosting regression tree, or random forest). Experiments show our framework performs well in all their combinations. 
Third, it is highly generalizable to other open search spaces, e.g. given a limited sample budget, we achieve the state-of-the-art ImageNet performance on the NASNet and MobileNet search spaces. Detailed comparison with state-of-the-art predictor-based NAS \citep{shi2019bridging,luo2020semi,wang2019sampleefficient,chau2020brp} is presented in Section 4.

\begin{figure*}[t]
\begin{center}
\vspace{-3em}
\includegraphics[width=1.0\textwidth]{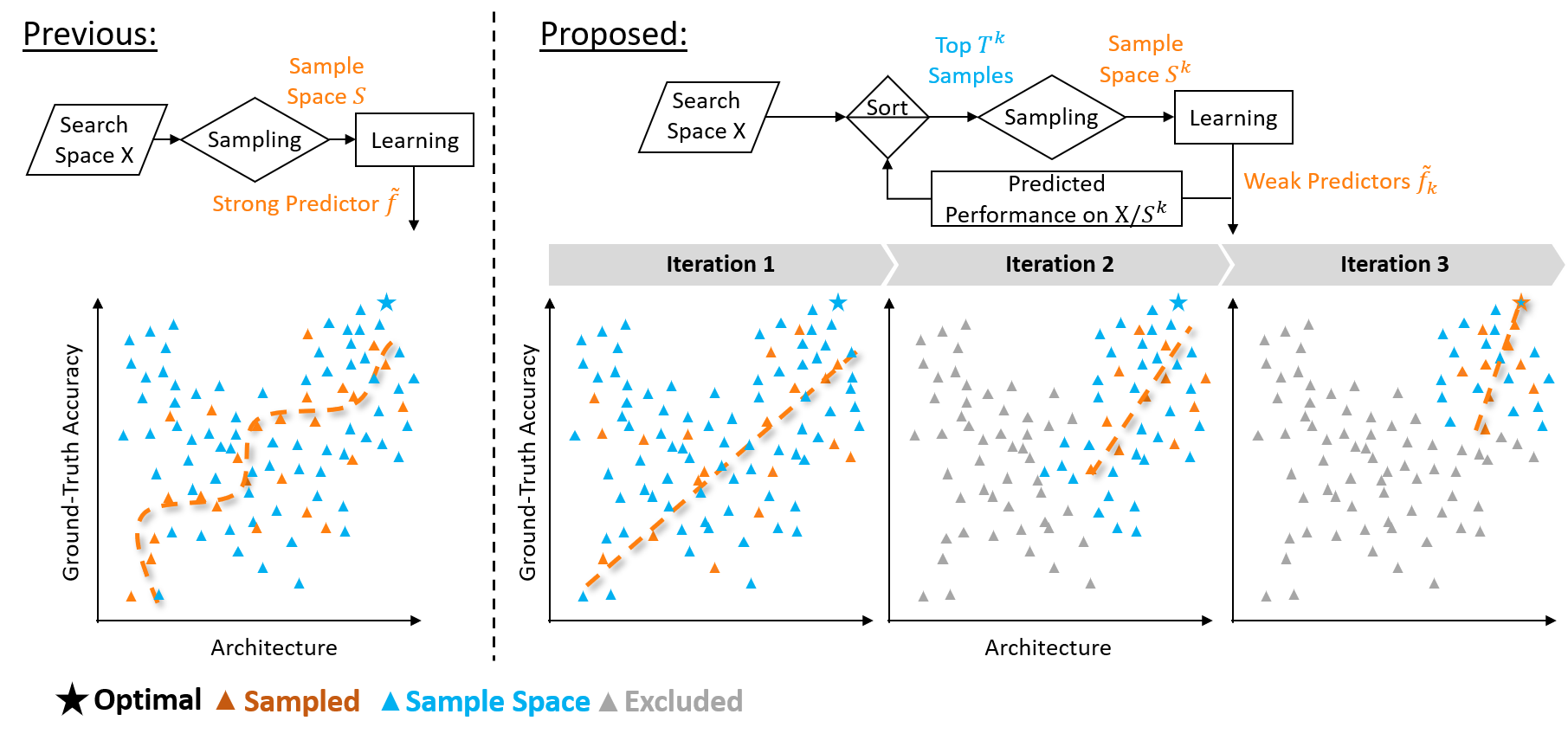}

\vspace{-1em}
\caption{An illustration of WeakNAS's progressive approximation. Previous predictor-based NAS uniformly sample in the whole search space to fit a strong predictor. Instead, our method progressively shrinks the sample space based on predictions from previous weak predictors, and update new weak predictors towards subspace of better architectures, hence focusing on fitting the search path.}
\vspace{-2em}
\label{fig:progressive}
\end{center}
\end{figure*}

\vspace{-0.5em}
\section{Our Framework}
\vspace{-0.5em}
\subsection{Reformulating Predictor-based NAS as Bi-Level Optimization}
\vspace{-0.5em}
Given a search space of network architectures $X$ and an architecture-to-performance mapping function $f:X\rightarrow P$ from the architecture set $X$ to the performance set $P$, the objective is to find the best neural architecture $x^{*}$ with the highest performance $f(x)$ in the search space $X$:
\begin{equation}
\begin{aligned}
& x^{*} = \argmax_{x \in X} f(x)
\label{eq:nas}
\end{aligned}
\end{equation}
A na{\"i}ve solution is to estimate the performance mapping $f(x)$ through the full search space. However, this is prohibitively expensive since all architectures have to be exhaustively trained from scratch. To address this problem, predictor-based NAS learns a proxy predictor $\tilde{f}(x)$ to approximate $f(x)$ by using some architecture-performance pairs, which significantly reduces the training cost. In general, predictor-based NAS can be re-cast as a bi-level optimization problem:
\begin{align}
\begin{split}
\quad x^{*} = \argmax_{x \in X} \tilde{f}(x | S), \,\,
\textrm{s.t.} \, \tilde{f} = \argmin_{S, \tilde{f} \in \tilde{\mathcal{F}}} \sum_{s \in S}\mathcal{L}(\tilde{f}(s), f(s))\,
\end{split}
\label{eq:nao}
\vspace{-0.5em}
\end{align}
where $\mathcal{L}$ is the loss function for the predictor $\tilde{f}$, $\tilde{\mathcal{F}}$ is a set of all possible approximation to $f$, $S:=\{S \subseteq X \mid  |S| \leq C\}$ all architectures satisfying the sampling budget $C$. $C$ is directly related to the total training cost, e.g., the total number of queries. Our objective is to minimize the loss $\mathcal{L}$ based on some sampled architectures $S$.

Previous predictor-based NAS methods attempt to solve Equation \ref{eq:nao} with two sequential steps: (1) \emph{sampling} some architecture-performance pairs and (2) \emph{learning} a proxy accuracy predictor. For the first step, a common practice is to sample training pairs $S$ uniformly from the search space $X$ to fit the predictor. Such sampling is however inefficient considering that the goal of NAS is only to find well-performed architectures without caring for the bad ones.

\vspace{-0.7em}
\subsection{Progressive Weak Predictors Emerge Naturally as A Solution to the Optimization}
\vspace{-0.5em}

\textbf{Optimization Insight:} Instead of first (uniformly) sampling the whole space and then fitting the predictor, we propose to jointly evolve the sampling $S$ and fit the predictor $\tilde{f}$, which helps achieve better sample efficiency by focusing on only relevant sample subspaces. That could be mathematically formulated as solving Equation \ref{eq:nao} in a new coordinate descent way, that iterates between optimizing the architecture \emph{sampling} and predictor \emph{fitting} subproblems:\\
\begin{equation}
\vspace{-1em}
\begin{aligned}
\vspace{-1em}
\textrm{(Sampling)} \quad
&\tilde{P}^{k} = \{\tilde{f}_{k}(s) \vert s \in X\setminus S^{k} \},\,\,
S_M \subset \text{Top}_N(\tilde{P}^{k}), \,\,
S^{k+1} = S_{M} \cup S^{k}, \,\,\\
&\text{where}\,\,\text{Top}_N(\tilde{P}^{k})\,\,\text{denote the set of top N architectures in}\,\,\tilde{P}^{k}
\end{aligned}
\end{equation}
\begin{equation}
\label{eq:weak-nas}
\begin{aligned}
\textrm{(Predictor Fitting)} \quad x^{*}= \argmax_{x \in X} \tilde{f}(x | S^{k+1}),\,\,
\textrm{s.t.}\,\, \tilde{f}_{k+1} = \argmin_{\tilde{f}_{k} \in \tilde{\mathcal{F}}} \sum_{s \in S^{k+1}}{\mathcal{L}}(\tilde{f}(s), f(s))
\vspace{-0.9em}
\end{aligned}
\end{equation}
In comparison, existing predictor-based NAS methods could be viewed as running the above coordinate descent \textit{for just one iteration} -- a special case of our general framework.

As well known in optimization, many iterative algorithms only need to solve (subsets of) their subproblems inexactly \citep{tappenden2016inexact,schmidt2011convergence,hager2020convergence} for properly ensuring convergence either theoretically or empirically. Here, using a strong predictor to fit the whole space could be treated as solving the predictor fitting subproblem relatively precisely, while adopting a weak predictor just imprecisely solves that. Previous methods solving Equation \ref{eq:nao} truncate their solutions to ``one shot" and hinge on solving subproblems with higher precision. Since we now take a joint optimization view and allow for multiple iterations, we can afford to only use weaker predictors for the fitting subproblem per iteration.

\textbf{Implementation Outline:} The above coordinate descent solution has clear interpretations and is straightforward to implement. Suppose our iterative methods has $K$ iterations. We initialize $S^{1}$ by randomly sampling a few samples from $X$, and train an initial predictor $\tilde{f}_{1}$. Then at iterations $k = 2, \dots K$, we jointly optimize the sampling set $S^{k}$ and predictor $\tilde{f}_{k}$ in an alternative manner.

\begin{figure*}[!htp]
\vspace{-2em}
\centering
  \begin{subfigure}[b]{0.314\textwidth} 
    \centering
    \includegraphics[width=\textwidth]{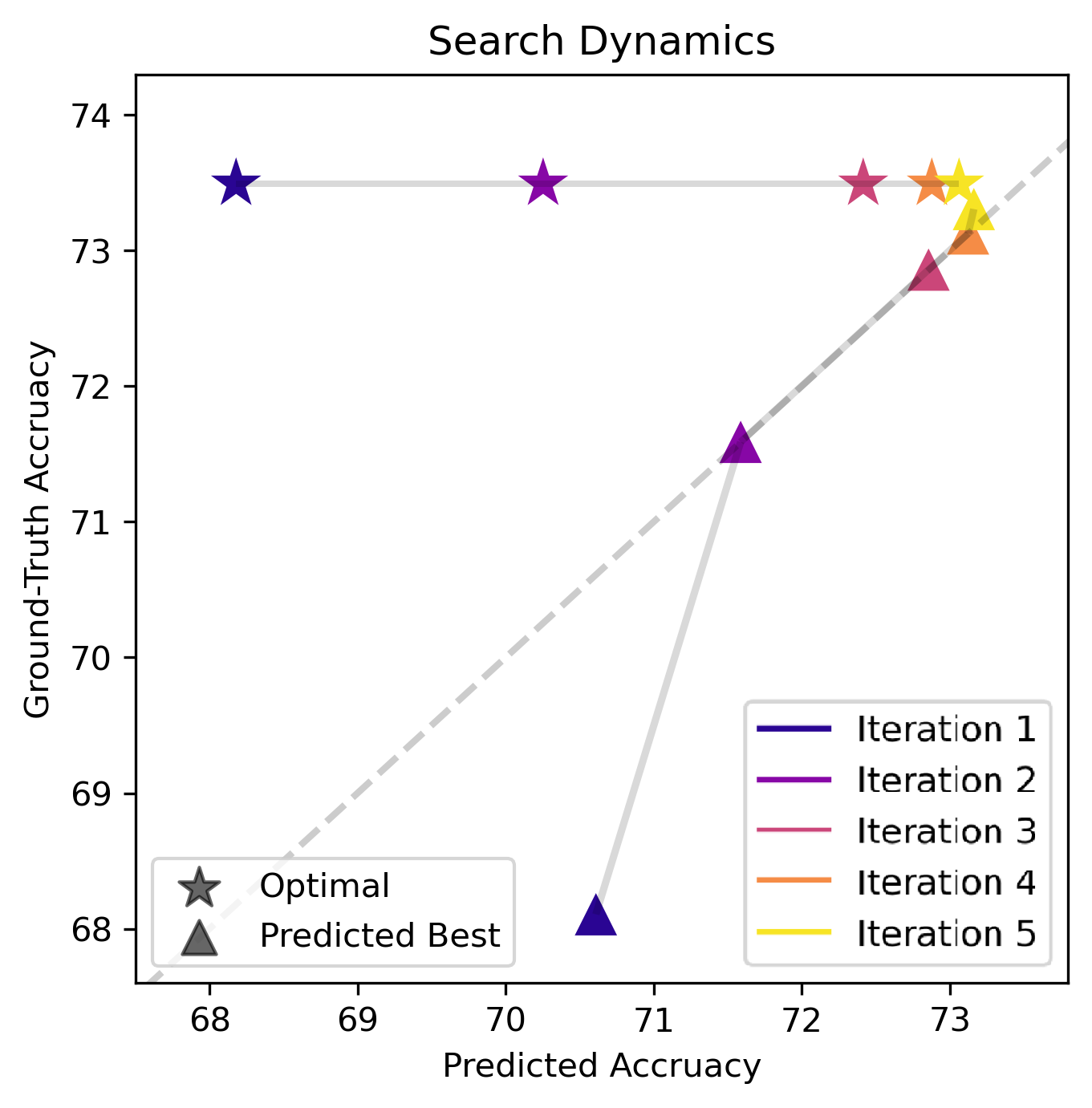}
    \vspace{-1.5em}
    \caption{}
  \end{subfigure}
  \hspace{-0.2em}\begin{subfigure}[b]{0.323\textwidth} 
    \centering
    \includegraphics[width=\textwidth]{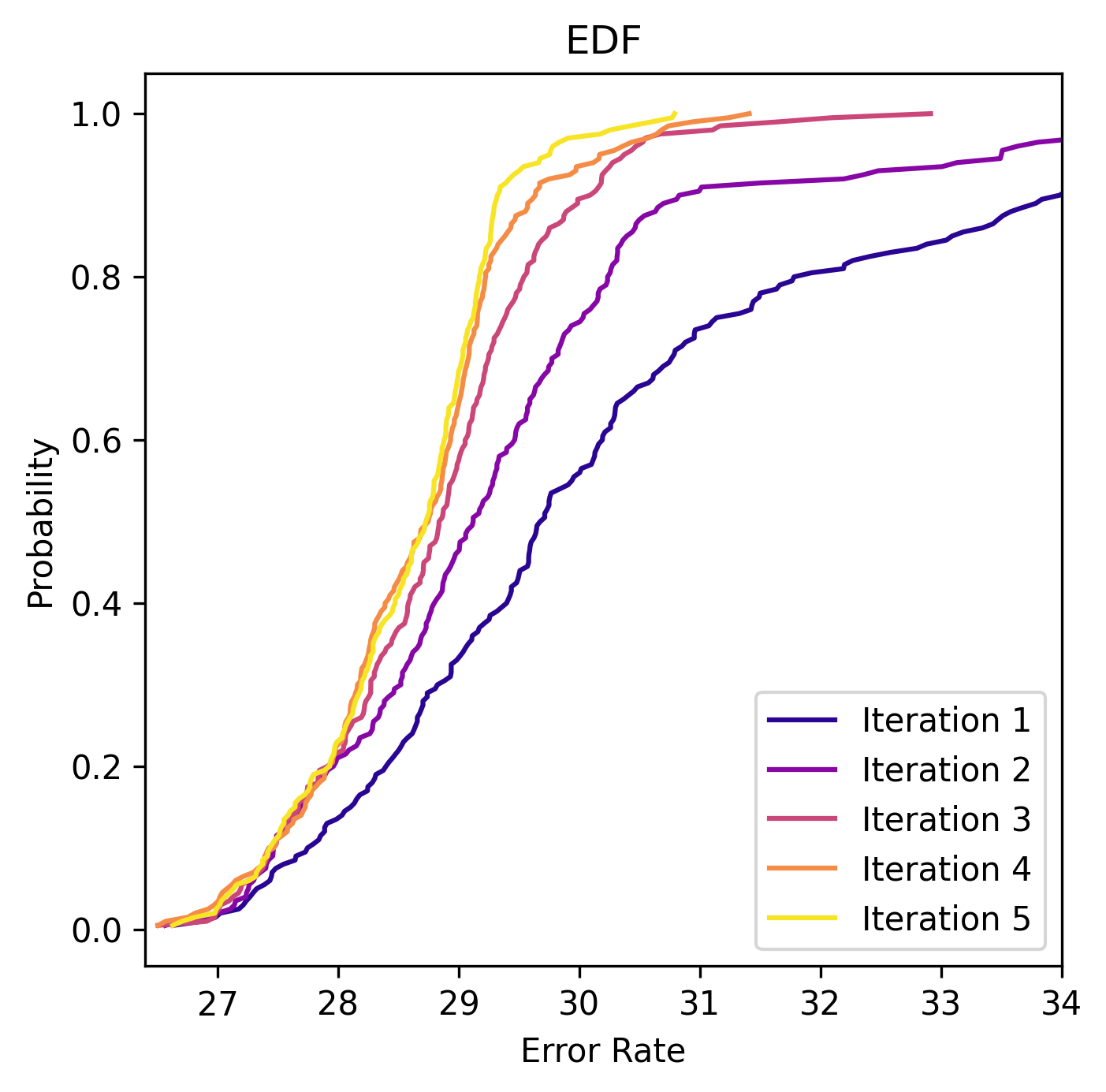}
    \vspace{-1.5em}
    \caption{}
  \end{subfigure}
  \hspace{-0.2em}\begin{subfigure}[b]{0.36\textwidth} 
    \centering
    \includegraphics[width=\textwidth]{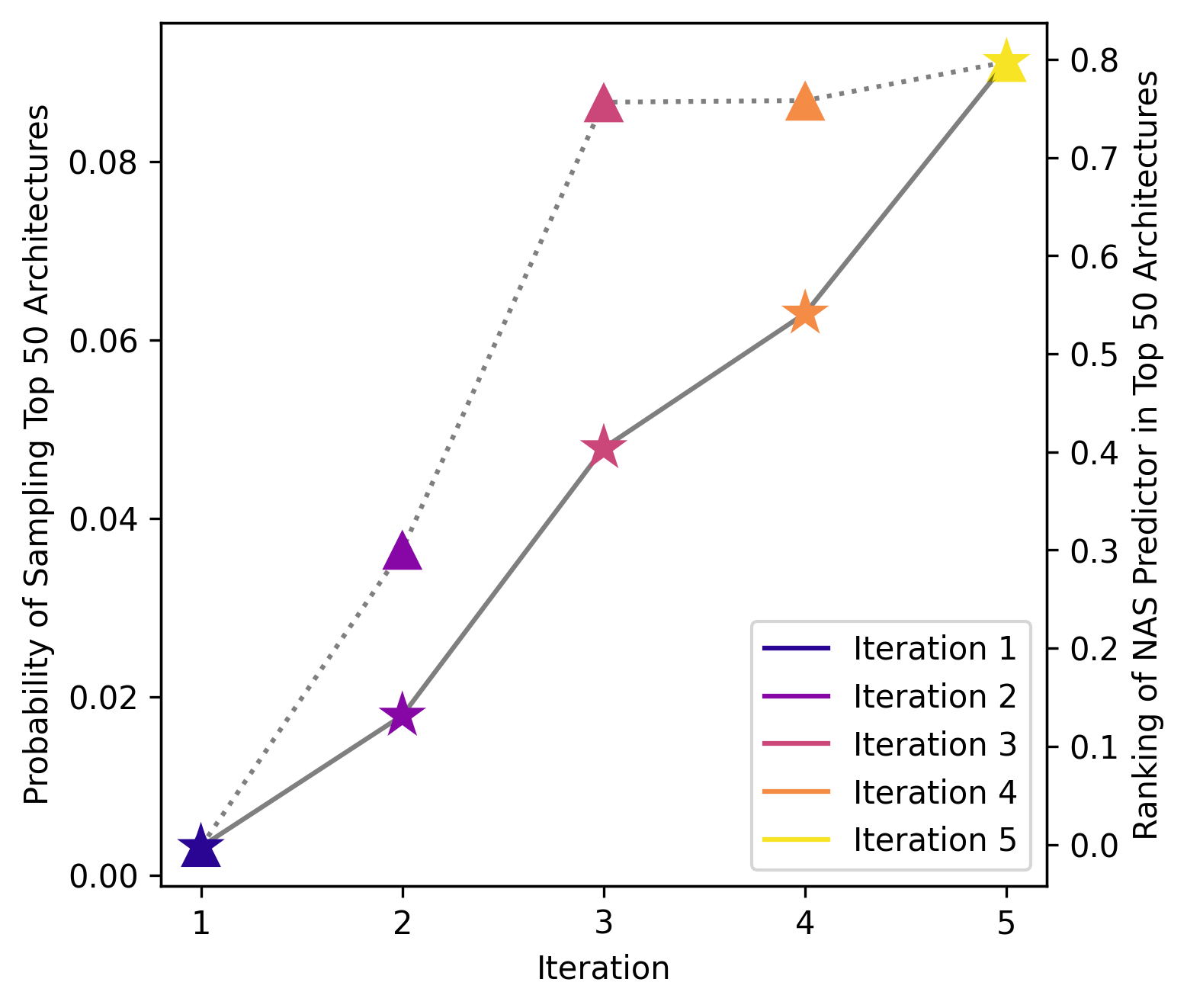}
    \vspace{-1.5em}
    \caption{}
  \end{subfigure}
\vspace{-1.5em}
\caption{Visualization of the search dynamics in NAS-Bench-201 Search Space. (best viewed in color) (a) The trajectory of the predicted best architecture and global optimal through out 5 iterations; (b) Error \textit{empirical distribution function} (EDF) of the predicted top-200 architectures throughout 5 iterations (c) Triangle marker: probability of sampling top-50 architectures throughout 5 iterations; Star marker: Kendall's Tau ranking of NAS predictor in Top 50 architectures through out 5 iterations.}
\label{fig:search_dynamics}
\end{figure*}

\begin{figure*}[t]
\vspace{-1.3em}
\begin{center}
\includegraphics[width=1.0\textwidth]{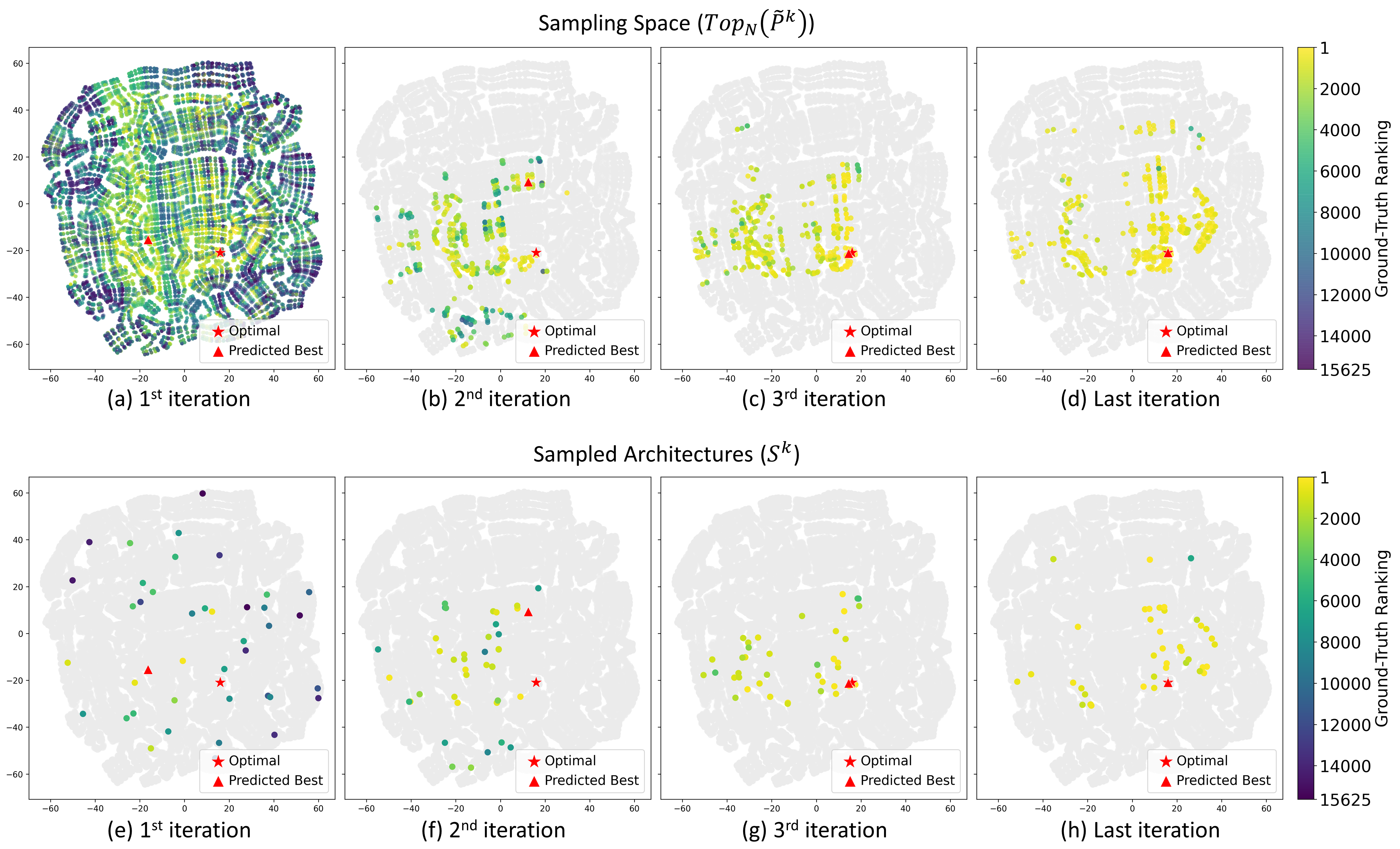}
\vspace{-1.2em}
\caption{Visualization of search dynamics in NAS-Bench-201 Search Space via t-SNE. At $i$-th iteration, we randomly sample M = 40 new architectures from the top N = 400 ranked architectures in $\tilde{P}^{k}$. The top row from (a)-(d) show the sampling space ${Top}_N(\tilde{P}^{k})$, and the bottom row from (e)-(h) show the sampled architectures ${S}^{k}$. The performance ranking of architectures is encoded by color, and those not-sampled architectures are colored in grey.}
\vspace{-2em}
\label{fig:tsne}
\end{center}
\end{figure*}

\vspace{-0.5em}
\paragraph{\textit{Subproblem 1: Architecture Sampling}.} At iteration $k+1$, we first sort all architectures\footnote{One only exception is the Section~\ref{sec:opendomainexp} open-domain experiments: we will sub-sample all architectures in the search space before sorting. More details can be found in Appendix Section~\ref{opendomaindetails}} in the search space $X$ (excluding all the samples already in $S^k$) according to its predicted performance $\tilde{P}^{k}$ at every iteration $k$. We then randomly sample $M$ new architectures from the top $N$ ranked architectures in $\tilde{P}^{k}$. Note this step both reduces the sample budget, and controls the exploitation-exploration trade-off (see Section \ref{sec:exploitation-exploration}). The newly sampled architectures together with $S^k$ become $S^{k+1}$. 

\vspace{-0.5em}
\paragraph{\textit{Subproblem 2: (Weak) Predictor Fitting}.} We learn a predictor $\tilde{f}^{k+1}$, by minimizing the loss $\mathcal{L}$ of the predictor $\tilde{f}^{k+1}$ based on sampled architectures $S^{k+1}$. We then evaluate architectures using the learned predictor $\tilde{f}^{k+1}$ to get the predicted performance $\tilde{P}^{k+1}$.

As illustrated in Figure \ref{fig:progressive}, through alternating iterations, 
we progressively evolve weak predictors to focus on sampling along the search path, thus simplifying the learning workload of each predictor. With these coarse-to-fine iterations, the predictor $\tilde{f}^{k}$ would guide the sampling process to gradually zoom into the promising architecture samples. In addition, the promising samples $S^{k+1}$ would in turn improve the performance of the updated predictor $\tilde{f}^{k+1}$ among the well-performed architectures, hence the ranking of sampling space is also refined gradually. In other words, the solution quality to the subproblem 2 will gradually increase as a natural consequence of the guided zoom-in. 
For derivation, we simply choose the best architecture predicted by the final weak predictor. This idea is related to the classical ensembling \citep{zhou2012ensemble}, yet a new regime to NAS.

\vspace{-1em}
\paragraph{\textit{Proof-of-Concept Experiment}.}Figure \ref{fig:search_dynamics} (a) shows the progressive procedure of finding the optimal architecture ${x}^{*}$ and learning the predicted best architecture $\tilde{x}_{k}^{*}$ over $5$ iterations. As we can see from Figure \ref{fig:search_dynamics} (a), the optimal architecture and the predicted best one are moving towards each other closer and closer, which indicates that the performance of predictor over the optimal architecture(s) is growing better.
In Figure \ref{fig:search_dynamics} (b), we use the error \textit{empirical distribution function} (EDF) \citep{radosavovic2020designing} to visualize the performance distribution of architectures in the subspace. We plot the EDF of the top-$200$ models based on the predicted performance over $5$ iterations. As is shown, the subspace of top-performed architectures is consistently evolving towards more promising architecture samples over $5$ iterations. Then in Figure \ref{fig:search_dynamics} (c), we validate that the probabilities of sampling better architectures within the top $N$ predictions keep increasing. Based on this property, we can just sample a few well-performing architectures guided by the predictive model to estimate another better weak predictor. The same plot also suggests that the NAS predictor's ranking among the top-performed models is gradually refined, since more and more architectures in the top region are sampled.

In Figure \ref{fig:tsne}, we also show the t-SNE visualization of the search dynamic in NAS-Bench-201 search space. We can observe that: (1) NAS-Bench-201 search space is highly structured; (2) the sampling space ${Top}_N(\tilde{P}^{k})$ and sampled architectures ${S}^{k}$ are both consistently evolving towards more promising regions, as can be noticed by the increasingly warmer color trend.

\vspace{-0.8em}
\subsection{Relationship to Bayesian Optimization: A Simplification and Why It Works}\label{sec:bo_relation}
\vspace{-0.5em}
Our method can be alternatively regarded as a \textbf{vastly simplified variant} of Bayesian Optimization (BO). It does not refer to any explicit uncertainty-based modeling such as Gaussian Process (which are often difficult to scale up); instead it adopts a \textit{very simple step function} as our acquisition function.
For a sample $x$ in the search space $X$, our special ``acquisition function" can be written as:
\begin{equation}
acq(x)=u(x-\theta)\cdot \epsilon 
\label{eq:acq}
\end{equation}
where the step function $u(x)$ is 1 if  $x \geq \theta$, and 0 otherwise;
$\epsilon$ is a random variable from the uniform distribution $U(0, 1)$; and $\theta$ is the threshold to split Top$N$ from the rest, according to their predicted performance $\tilde{P}^{k}(x)$. We then choose the samples with the $M$ largest acquisition values:
\begin{equation}
S_{M} = \argmax_{\text{Top}M} acq(x)
\end{equation}
\textit{Why such ``oversimplified BO" can be effectively for our framework?} We consider the reason to be the inherently structured NAS search space. Specifically, existing NAS spaces are created either by varying operators from a pre-defined operator set (DARTS/NAS-Bench-101/201 Search Space) or by varying kernel size, width or depth (MobileNet Search Space). Therefore, as shown in Figure \ref{fig:tsne}, the search spaces are often highly-structured, and the best performers gather close to each other.

Here comes our underlying prior assumption: \textit{we can progressively connect a piecewise search path from the initialization, to the finest subspace where the best architecture resides.} At the beginning, since the weak predictor only roughly fits the whole space, the sampling operation will be ``noisier", inducing more exploration. When it comes to the later stage, the weak predictors fit better on the current well-performing clusters, thus performing more exploitation locally. Therefore our progressive weak predictor framework provides a natural evolution between exploration and exploitation, without explicit uncertainty modeling, thanks to the prior of special NAS space structure.

Another exploration-exploitation trade-off is implicitly built in the adaptive sampling step of our subproblem 1 solution. To recall, at each iteration, instead of choosing all Top $N$ models by the latest predictor, we randomly sample $M$ models from Top $N$ models to explore new architectures in a stochastic manner. By varying the ratio $\epsilon=M/N$ and the sampling strategy (e.g., uniform, linear-decay or exponential-decay), we can balance the sampling exploitation and exploration per step, in a similar flavor to the $\epsilon$-greedy \citep{sutton2018reinforcement} approach in reinforcement learning.

\vspace{-0.8em}
\subsection{Our Framework is General to Predictor Models and Architecture Representations}
\vspace{-0.5em}
\label{sec:generalize}

\begin{figure*}[!htp]
\vspace{-0.5em}
\begin{minipage}[]{1.0\textwidth}

\hspace{-1em}\subfloat[CIFAR10]{\includegraphics[width=0.348\textwidth]{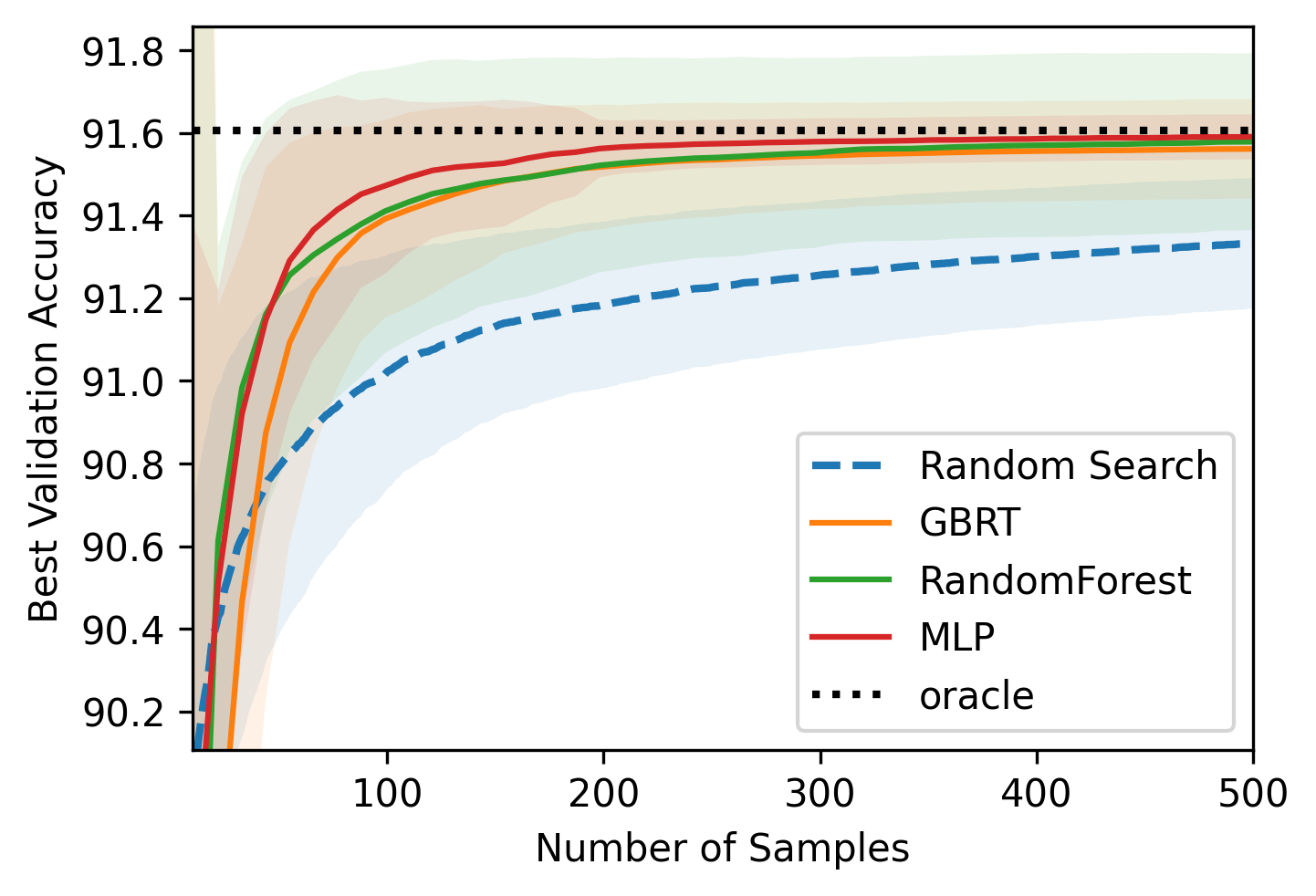}\vspace{-0.5em}}
\subfloat[CIFAR100]{\includegraphics[width=0.34\textwidth]{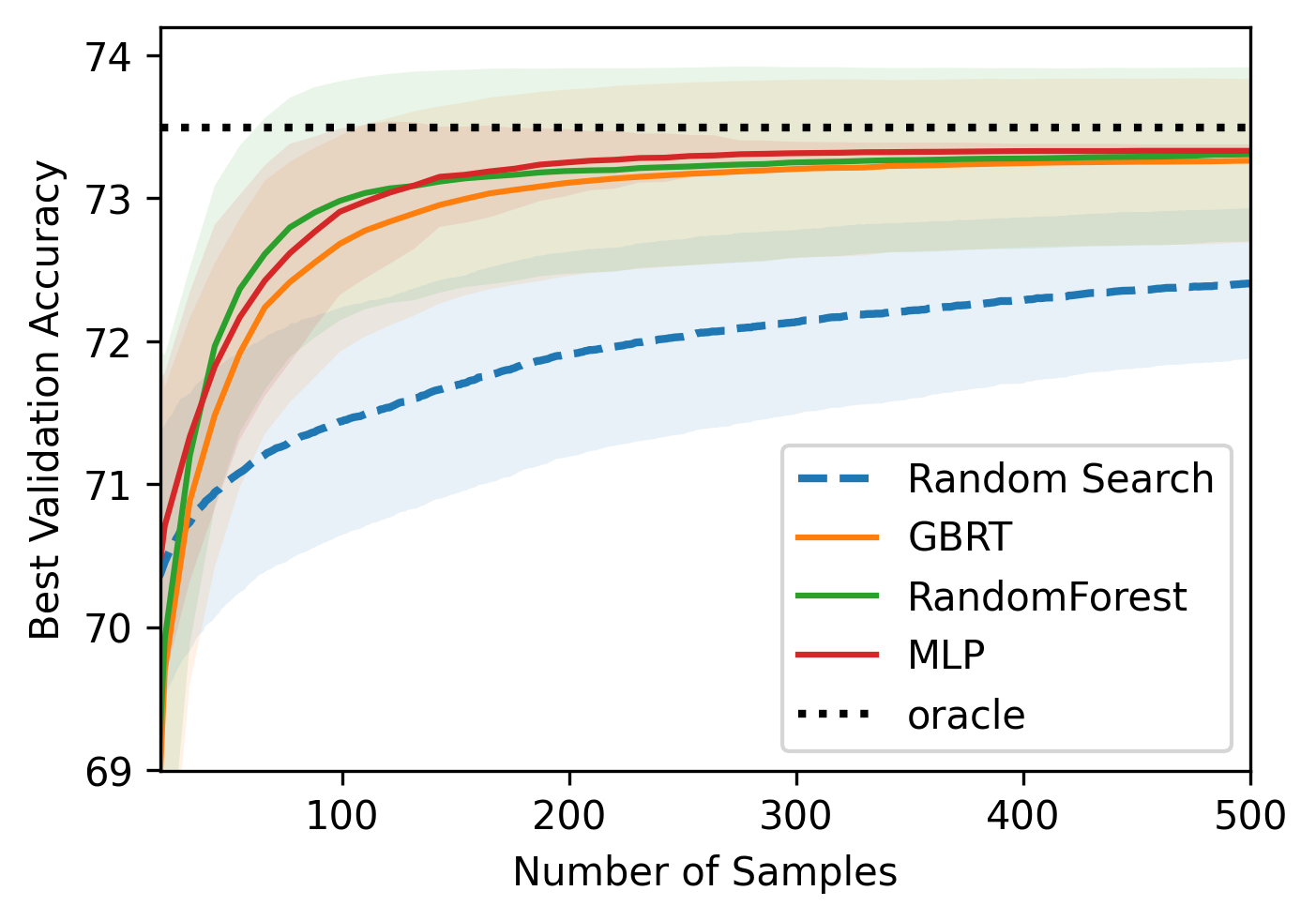}\vspace{-0.5em}}
\subfloat[ImageNet16-120]{\includegraphics[width=0.348\textwidth]{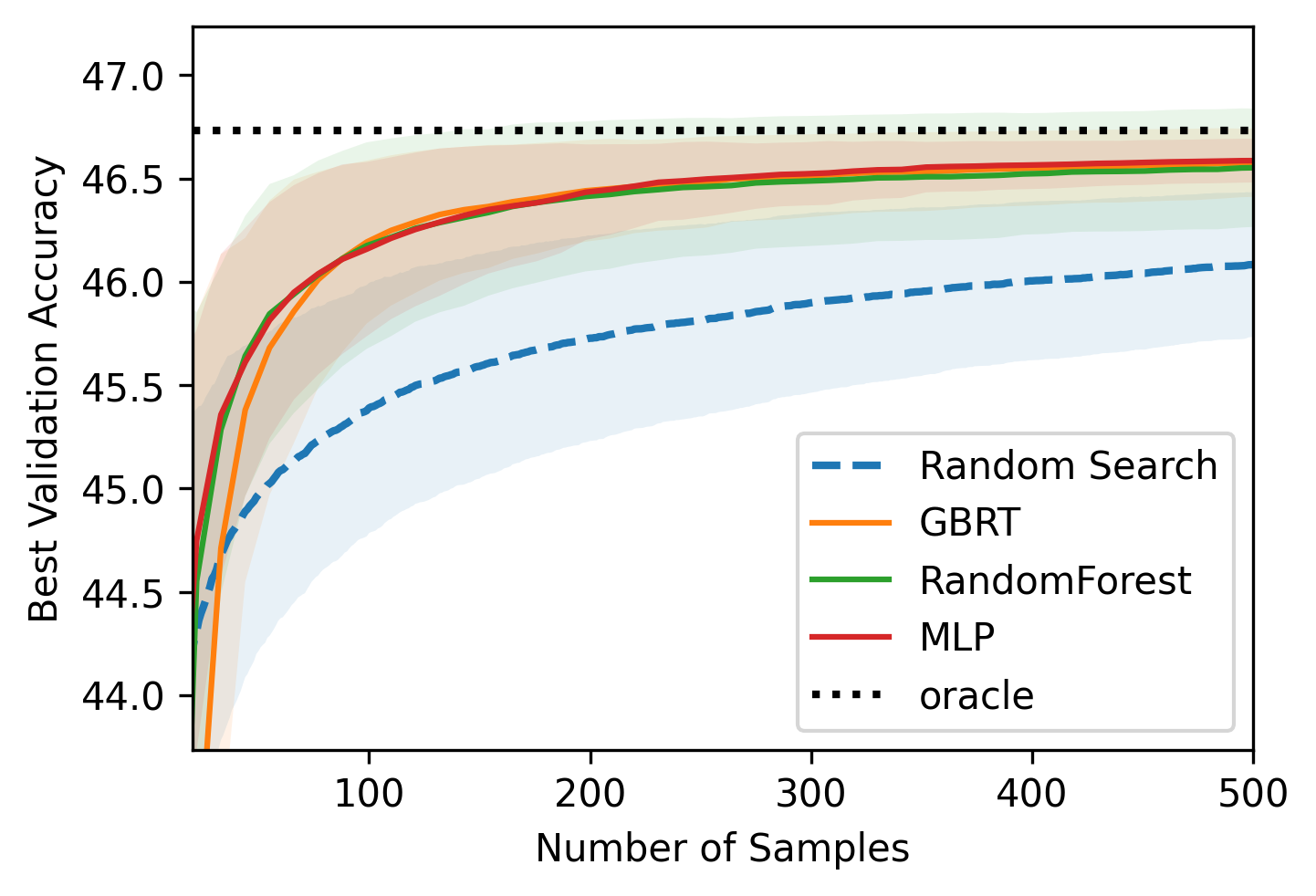}\vspace{-0.5em}}\\
\caption{Evaluations of robustness across different predictors on NAS-Bench-201. Solid lines and shadow regions denote the mean and std, respectively.}
\label{fig:robust}
\end{minipage}
\vspace{-1em}
\end{figure*}

Our framework is designed to be generalizable to various predictors and features. In predictor-based NAS, the objective of fitting the predictor $\tilde{f}$ is often cast as a regression~\citep{wen2019neural} or ranking~\citep{ning2020generic} problem. The choice of predictors is diverse, and usually critical to final performance \citep{ning2020generic, wei2020npenas, luo2018neural, wen2019neural, chau2020brp, luo2020neural}. To illustrate our framework is generalizable and robust to the specific choice of predictors, we compare the following predictor variants. 
\begin{itemize}

    \vspace{-0.3em}
    \item \textit{Multilayer perceptron (MLP)}: MLP is the common baseline in predictor-based NAS \citep{ning2020generic} due to its simplicity. For our weak predictor, we use a 4-layer MLP with hidden layer dimension of (1000, 1000, 1000, 1000).
    \vspace{-0.2em}
    \item \textit{Regression Tree}: tree-based methods are also popular \citep{luo2020neural, siems2020bench} since they are suitable for categorical architecture representations. As our weak predictor, we use the Gradient Boosting Regression Tree  (GBRT) based on XGBoost \citep{chen2016xgboost}, consisting of 1000 Trees.
    \vspace{-0.2em}
    \item \textit{Random Forest}: random forests differ from GBRT in that they combines decisions only at the end rather than along the hierarchy, and are often more robust to noise. For each weak predictor, we use a random forest consisting of 1000 Forests.
    \vspace{-0.3em}
\end{itemize}
The features representations to encode the architectures are also instrumental. Previous methods hand-craft various features for the best performance, e.g., raw architecture encoding \citep{wei2020npenas}, supernet statistics \citep{hu2020angle}, and
graph convolutional network encoding \citep{wen2019neural,ning2020generic,chau2020brp,shi2019bridging}
Our framework is also agnostic to various architecture representations, and we compare the following:
\vspace{-0.3em}
\begin{itemize}
    \item \textit{One-hot vector}: In NAS-Bench-201 \citep{Dong2020NAS-Bench-201:}, its DARTS-style search space has fixed  graph connectivity, hence the one-hot vector is commonly used to encode the choice of operator.
    \vspace{-0.2em}
    \item \textit{Adjacency matrix}: In NAS-Bench-101, we used the same encoding scheme as in \citep{ying2019bench, wei2020npenas}, where a 7$\times$7 adjacency matrix represents the graph connectivity and a 7-dimensional vector represents the choice of operator on every node.
    \vspace{-0.3em}
\end{itemize}
As shown in Figure \ref{fig:robust}, all predictor models perform similarly across different datasets. Comparing performance on NAS-Bench-101 and NAS-Bench-201, although they use different architecture encoding methods, our method still performs similarly well among different predictors. This demonstrates that our framework is robust to various predictor and feature choices.

\vspace{-1.0em}

\section{Experiments}
\vspace{-0.7em}

\label{sec:setup}

\noindent\textbf{Setup:} For all experiments, we use an Intel Xeon E5-2650v4 CPU and a single Tesla P100 GPU, and use the Multilayer perceptron (MLP) as our default NAS predictor, unless otherwise specified.

\noindent\textbf{NAS-Bench-101} \citep{ying2019bench} provides a Directed Acyclic Graph (DAG) based cell structure. The connectivity of DAG can be arbitrary with a maximum number of 7 nodes and 9 edges. Each nodes on the DAG can choose from operator of 1$\times$1 convolution, 3$\times$3 convolution or 3$\times$3 max-pooling. After removing duplicates, the dataset consists of 423,624 diverse architectures trained on CIFAR10\citep{krizhevsky2009learning}.

\noindent\textbf{NAS-Bench-201} \citep{Dong2020NAS-Bench-201:} is a more recent benchmark with a reduced DARTS-like search space. The DAG of each cell is fixed, and one can choose from 5 different operations (1$\times$1 convolution, 3$\times$3 convolution, 3$\times$3 avg-pooling, skip, no connection), on each of the 6 edges, totaling 15,625 architectures. It is trained on 3 different datasets:  CIFAR10, CIFAR100 and ImageNet16-120 \citep{chrabaszcz2017downsampled}. For experiments on both NAS-Benches, we followed the same setting as \cite{chau2020brp}.

\noindent\textbf{Open Domain Search Space:} We follow the same NASNet search space used in \citep{zoph2018learning} and MobileNet Search Space used in \citep{cai2019once} to directly search for the best architectures on ImageNet\citep{deng2009imagenet}. Due to the huge computational cost to evaluate sampled architectures on ImageNet, we leverage a weight-sharing supernet approach. On NASNet search space, we use Single-Path One-shot \citep{guo2019single} approach to train our SuperNet, while on MobileNet Search Space we reused the pre-trained supernet from OFA\citep{cai2019once}.
We then use the supernet accuracy as the performance proxy to train weak predictors. We clarify that despite using supernet, our method is more accurate than existing differentiable weight-sharing methods, meanwhile requiring less samples than evolution based weight-sharing methods, as manifested in Table \ref{table:imagenet-nasnet} and  \ref{table:imagenet-mobilenet}. We adopt PyTorch and image models library (timm) \citep{rw2019timm} to implement our models and conduct all ImageNet experiments using 8 Tesla V100 GPUs. For derived architecture, we follow a similar training from scratch strategies used in LaNAS\citep{wang2019sampleefficient}.

\vspace{-0.8em}
\subsection{Ablation Studies}
\label{sec:exploitation-exploration}
\vspace{-0.5em}
We conduct a series of ablation studies on the effectiveness of proposed method on NAS-Bench-101. To validate the effectiveness of our iterative scheme, In Table \ref{table:progressive}, we initialize the initial Weak Predictor $\tilde{f}_{1}$ with 100 random samples, and set $M=10$, after progressively adding more weak predictors (from 1 to 191), we find the performance keeps growing. This demonstrates the key property of our method that probability of sampling better architectures keeps increasing as more iteration goes.
It's worth noting that the quality of random initial samples $M_{0}$ may also impact on the performance of WeakNAS, but if $|M_{0}|$ is sufficiently large, the chance of hitting good samples (or its neighborhood) is high, and empirically we found $|M_{0}|$=100 to already ensure highly stable performance at NAS-Bench-101: a more detailed ablation can be found in the Appendix Section~\ref{sec:num_init}.

\begin{table*}[!htp]
\vspace{-0.5em}
\centering
\scalebox{0.92}{\begin{tabular}{lrrcccr}
\toprule
Sampling & \#Predictor & \#Queries &  Test Acc.(\%) & SD(\%) & Test Regret(\%) & Avg. Rank\\
\midrule
Uniform & 1 Strong Predictor & 2000 & 93.92 & 0.08 & 0.40 & 135.0 \\
\midrule

\multirow{5}{*}{Iterative} & 1 Weak Predictor & 100 & 93.42 & 0.37 & 0.90 & 6652.1 \\
& 11 Weak Predictors & 200 & 94.18 & 0.14 & 0.14 & 5.6 \\
& 91 Weak Predictors & 1000  & 94.25 & 0.04 & 0.07 & 1.7 \\
& 191 Weak Predictors & 2000 & 94.26 & 0.04 & 0.06 & 1.6 \\
\midrule
Optimal & - & - & 94.32 & - & 0.00 & 1 \\
\bottomrule
\end{tabular}}
\caption{Ablation on the effectiveness of our iterative scheme on NAS-Bench-101} 
\label{table:progressive}
\vspace{-0.5em}
\end{table*}

\begin{table*}[!htp]
\vspace{-0.5em}
\centering
\scalebox{0.88}{\begin{tabular}{lcrccccr}
\toprule
 Sampling (M from TopN) & M & TopN &\#Queries &  Test Acc.(\%) & SD(\%) & Test Regret(\%) & Avg. Rank\\
\midrule
Exponential-decay & 10 & 100 & 1000 & 93.96 & 0.10 & 0.36 & 85.0 \\
Linear-decay & 10 & 100 & 1000 & 94.06 & 0.08 & 0.26 & 26.1 \\
\textbf{Uniform} & \textbf{10} & \textbf{100} & \textbf{1000} & \textbf{94.25}  & \textbf{0.04} & \textbf{0.07} & \textbf{1.7}\\
\midrule
Uniform & 10 & 1000 & 1000 & 94.10 & 0.19 & 0.22 & 14.1 \\
\textbf{Uniform} & \textbf{10} & \textbf{100} & \textbf{1000} & \textbf{94.25} & \textbf{0.04} & \textbf{0.07} & \textbf{1.7}\\
Uniform & 10 & 10 & 1000 & 94.24 & 0.04 & 0.08 & 1.9 \\
\bottomrule
\end{tabular}}
\caption{Ablation on exploitation-exploration trade-off on NAS-Bench-101}
\label{table:exploration}
\vspace{-0.3em}
\end{table*}

\begin{table*}[!htp]
\vspace{-1.0em}
\centering
\scalebox{0.8}{
\begin{tabular}{lrcccr}
\toprule
Method & \#Queries &  Test Acc.(\%) & SD(\%) & Test Regret(\%)  & Avg. Rank \\
\midrule
\textbf{WeakNAS} & \textbf{1000} & \textbf{94.25} & \textbf{0.04} & \textbf{0.07} & \textbf{1.7} \\
\midrule
WeakNAS (BO Variant) & 1000 & 94.12 & 0.15 & 0.20 & 8.7 \\
\midrule
Optimal & - & 94.32 & - & 0.00 & 1.0 \\
\bottomrule
\end{tabular}}
\caption{Comparing to the BO variant of WeakNAS on NAS-Bench-101.}
\label{table:bo-variant}

\end{table*}

\begin{wrapfigure}{r}{0.5\textwidth}
  \vspace{-0.3em}
  \begin{center}
    \vspace{-1.5em}
    \includegraphics[width=0.48\textwidth]{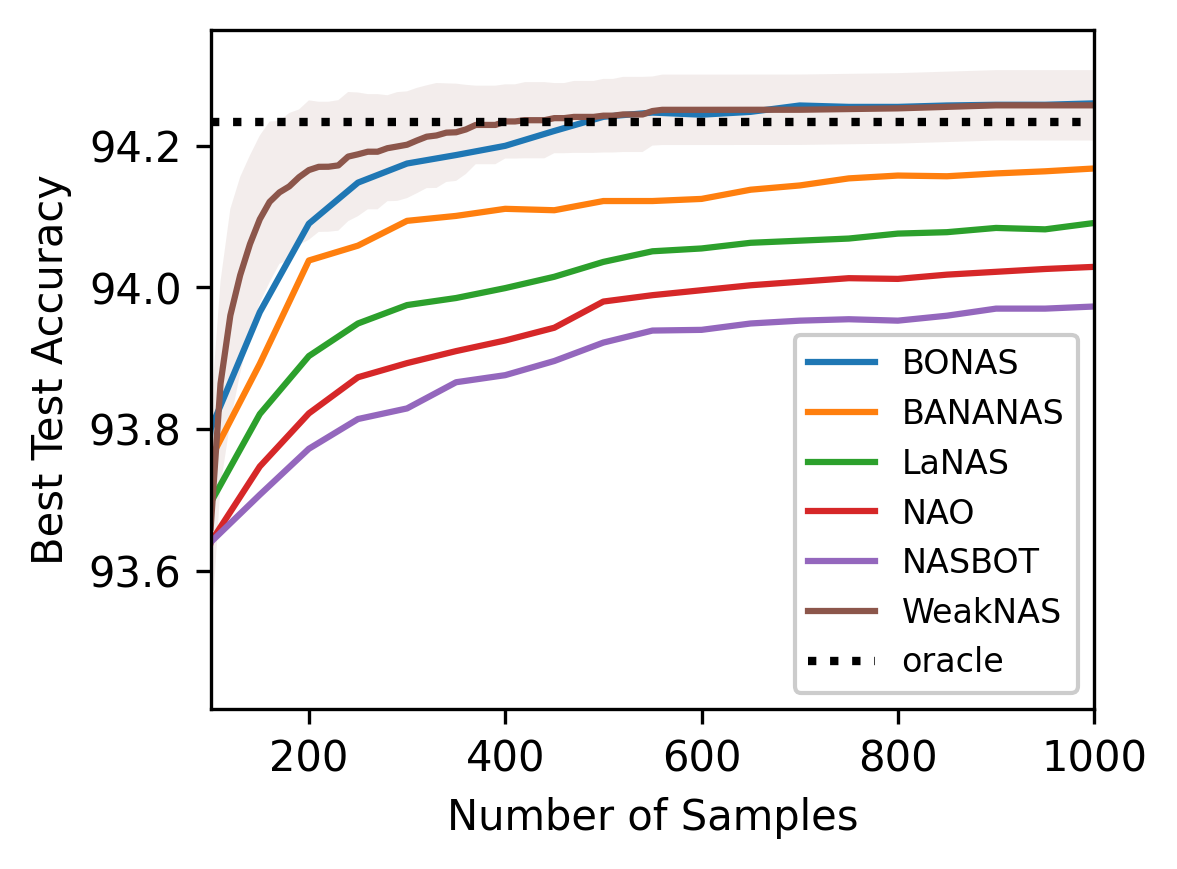}
    \vspace{-1.5em}
  \end{center}
  \caption{Comparison with SoTA methods on NAS-Bench-101. Solid lines and shadow regions denote the mean and std, respectively.}
  \vspace{-1.2em}
\label{fig:nasbench101-sota}
\end{wrapfigure}

We then study the exploitation-exploration trade-off in Table \ref{table:exploration} in NAS-Bench-101 (a similar ablation in Mobilenet Search space on ImageNet is also included in Appendix Table~\ref{table:exploration-mobilenet}) by investigating two settings: (a) We gradually increase $N$ to allow for more exploration, similar to controlling $\epsilon$ in the epsilon-greedy \citep{sutton2018reinforcement} approach in the RL context; (b) We vary the sampling strategy from Uniform, Linear-decay to Exponential-decay (top models get higher probabilities by following either linear-decay or exponential-decay distribution). We empirically observed that: (a) The performance drops more (Test Regret 0.22\% vs 0.08\%) when more exploration (TopN=1000 vs TopN=10) is used. This indicates that extensive exploration is not optimal for NAS-Bench-101; (b) Uniform sampling method yields better performance than sampling method that biased towards top performing model (e.g. linear-decay, exponential-decay). This indicates good architectures are evenly distributed within the Top 100 predictions of Weak NAS, therefore a simple uniform sampling strategy for exploration is more optimal in NAS-Bench-101. To conclude, our Weak NAS Predictor strikes a good balance between exploration and exploration.

Apart from the above exploitation-exploration trade-off of WeakNAS, we also explore the possibility of integrating other meta-sampling methods. We found that the local search algorithm could achieve comparable performance, while using Semi-NAS \citep{luo2020semi} as a meta sampling method could further boost the performance of WeakNAS: more details are in Appendix Section \ref{sec:meta-sampling}.

\vspace{-0.8em}
\subsection{Comparison to State-of-the-art (SOTA) Methods}
\vspace{-0.5em}

\noindent\textbf{NAS-Bench-101:} On NAS-Bench-101 benchmark, we compare our method with several popular methods \citep{real2019regularized, wang2019alphax, wang2019sampleefficient, luo2018neural, wen2019neural, luo2020semi, shi2019bridging, yan2020does, tang2020semi, yan2021cate, white2019bananas}.

Table \ref{table:optimal} shows that our method significantly outperforms baselines in terms of sample efficiency. Specifically, our method costs 964$\times$, 447$\times$, 378$\times$, 245$\times$, 58$\times$, and 7.5$\times$ less samples to reach the optimal architecture, compared to Random Search, Regularized Evolution \citep{real2019regularized}, MCTS \citep{wang2019alphax}, Semi-NAS\citep{luo2020semi}, LaNAS\citep{wang2019sampleefficient}, BONAS\citep{shi2019bridging}, respectively.
We then plot the best accuracy against number of samples in Table~\ref{table:sota-nasbench101} and Figure~\ref{fig:nasbench101-sota} to show the sample efficiency on the NAS-Bench-101, from which we can see that our method consistently costs fewer sample to reach higher accuracy.

\begin{table*}[!htp]
\vspace{-0.5em}
\centering
\scalebox{0.93}{
\begin{tabular}{lrcccr}
\toprule
Method & \#Queries &  Test Acc.(\%) & SD(\%) & Test Regret(\%)  & Avg. Rank \\
\midrule
Random Search & 2000 & 93.64 & 0.25 & 0.68 & 1750.0 \\
NAO \citep{luo2018neural} & 2000 & 93.90 & 0.03 & 0.42 & 168.1 \\
Reg Evolution \citep{real2019regularized} & 2000 & 93.96 & 0.05 & 0.36 & 85.0 \\
Semi-NAS \citep{luo2020semi} & 2000 & 94.02 & 0.05 & 0.30 & 42.1 \\
Neural Predictor \citep{wen2019neural} & 2000 & 94.04 & 0.05 & 0.28 & 33.5 \\
\multirow{1}{*}{\textbf{WeakNAS}} & \textbf{2000} & \textbf{94.26} & \textbf{0.04} & \textbf{0.06} & \textbf{1.6} \\
\midrule
Semi-Assessor \citep{tang2020semi} & 1000 & 94.01 & - & 0.31 & 47.1\\
LaNAS \citep{wang2019sampleefficient} & 1000 & 94.10 & - & 0.22 & 14.1 \\
BONAS \citep{shi2019bridging} & 1000 & 94.22 & - & 0.10 & 3.0\\
\multirow{1}{*}{\textbf{WeakNAS}} & \textbf{1000} & \textbf{94.25} & \textbf{0.04} & \textbf{0.07} & \textbf{1.7} \\ 
\midrule
Arch2vec \citep{yan2020does} & 400 & 94.10 & - & 0.22 & 14.1\\
\textbf{WeakNAS} & \textbf{400} & \textbf{94.24} & \textbf{0.04} & \textbf{0.08} & \textbf{1.9} \\ 
\midrule
LaNAS \citep{wang2019sampleefficient} & 200 & 93.90 & - & 0.42 & 168.1 \\
BONAS \citep{shi2019bridging} & 200 & 94.09 & - & 0.23 & 18.0 \\
\textbf{WeakNAS} & \textbf{200} & \textbf{94.18} & \textbf{0.14} & \textbf{0.14} & \textbf{5.6} \\ 
\midrule
NASBOWLr \cite{ru2020interpretable} & 150 & 94.09 & - & 0.23 & 18.0 \\ 
CATE (cate-DNGO-LS) \citep{yan2021cate} & 150 & 94.10 & - & 0.22 & 12.3 \\
\textbf{WeakNAS} & \textbf{150} & \textbf{94.10} & \textbf{0.19} & \textbf{0.22} & \textbf{12.3} \\
\midrule
Optimal & - & 94.32 & - & 0.00 & 1.0 \\
\bottomrule
\end{tabular}}
\caption{Comparing searching efficiency by limiting the total query amounts on NAS-Bench-101.}
\label{table:sota-nasbench101}
\vspace{-0.5em}
\end{table*}

\begin{table*}[!htp]
\centering
\vspace{-0.5em}
\setlength{\tabcolsep}{4.0mm}{
\scalebox{0.91}{
\begin{tabular}{c|r|rrr}
\toprule
Method & NAS-Bench-101 & \multicolumn{3}{c}{NAS-Bench-201}\\
\midrule
Dataset & CIFAR10 & CIFAR10 & CIFAR100 & ImageNet16-120\\
\midrule
Random Search 
 & 188139.8 & 7782.1 & 7621.2 & 7726.1 \\
Reg Evolution \citep{real2019regularized} 
& 87402.7 & 563.2 & 438.2 & 715.1 \\
MCTS \citep{wang2019alphax}
& 73977.2 & \textsuperscript{$\dagger$}528.3 & \textsuperscript{$\dagger$}405.4 & \textsuperscript{$\dagger$}578.2 \\
Semi-NAS \citep{luo2020semi} 
& \textsuperscript{$\dagger$}47932.3 & - & - & - \\
LaNAS \citep{wang2019sampleefficient}
& 11390.7 & \textsuperscript{$\dagger$}247.1\ & \textsuperscript{$\dagger$}187.5 & \textsuperscript{$\dagger$}292.4 \\
BONAS \citep{shi2019bridging}
& 1465.4 & - & - & - \\
\midrule
\textbf{WeakNAS}
& \textbf{195.2}  & \textbf{182.1} & \textbf{78.4} & \textbf{268.4} \\
\bottomrule
\end{tabular}}}
\caption{Comparison on the number of samples required to find the global optimal on NAS-Bench-101 and NAS-Bench-201. \textsuperscript{$\dagger$} denote reproduced results using adapted code.}
\label{table:optimal}
\vspace{-0.5em}
\end{table*}

\noindent\textbf{NAS-Bench-201:} We further evaluate on NAS-Bench-201, and compare with random search, Regularized Evolution \citep{real2019regularized}, Semi-NAS\citep{luo2020semi}, LaNAS\citep{wang2019sampleefficient}, BONAS\citep{shi2019bridging}. . As shown in Table \ref{table:optimal}, we conduct searches on all three subsets (CIFAR10, CIFAR100, ImageNet16-120) and report the average number of samples needed to reach global optimal  on the testing set over 100 runs. It shows that our method has the smallest sample cost among all settings.

\label{sec:opendomainexp}
\noindent\textbf{Open Domain Search:} we further apply our method to open domain search without ground-truth, and compare with several popular methods \citep{zoph2018learning, real2019regularized, liu2018progressive, luo2018neural, yu2020bignas, dai2020fbnetv3, wang2019sampleefficient}. As shown in Tables \ref{table:imagenet-nasnet} and \ref{table:imagenet-mobilenet}, using the fewest samples (and only a fraction of GPU hours) among all, our method can achieve state-of-the-art ImageNet top-1 accuracies with comparable parameters and FLOPs. Our searched architecture is also competitive to expert-design networks. On the NASNet Search Space, compared with the SoTA predictor-based NAS method LaNAS (Oneshot) \citep{wang2019sampleefficient}, our method reduces 0.6\% top-1 error while using less GPU hours. On the MobileNet Search Space, we improve the previous SoTA LaNAS \citep{wang2019sampleefficient} to 81.3\% top-1 accuracy on ImageNet while costing less FLOPs.

\vspace{-1.3em}
\subsection{Discussion: Further Comparison with SOTA Predictor-based NAS Methods}
\label{sec:compare}
\vspace{-0.5em}
\noindent\textbf{BO-based NAS methods~\cite{shi2019bridging, ru2020interpretable}}: BO-based methods in general treat NAS as a black-box optimization problem, for example, BONAS~\citep{shi2019bridging} customizes the classical BO framework in NAS with GCN embedding extractor and Bayesian Sigmoid Regression to acquire and select candidate architectures. The latest BO-based NAS approach, NASBOWL~\cite{ru2020interpretable}, combines the Weisfeiler-Lehman graph kernel in BO to capture the topological structures of the candidate architectures.

Compare with those BO-based method, our WeakNAS is an ``oversimplified" version of BO as explained in Section~\ref{sec:bo_relation}. Interestingly, results in Table~\ref{table:sota-nasbench101} suggests that WeakNAS is able to outperform BONAS~\citep{shi2019bridging}, and is comparable to NASBOWLr~\cite{ru2020interpretable} on NAS-Bench-101, showcasing that the simplification does not compromise NAS performance.
We hypothesize that the following factors might be relevant: (1) the posterior modeling and uncertainty estimation in BO might be noisy; (2) the inherently structured NAS search space (shown in Figure~\ref{fig:tsne}) could enable a ``shortcut" simplification to explore and exploit. In addition, the conventional uncertainty modeling in BO, such as the Gaussian Process used by \cite{ru2020interpretable}, is not as scalable when the number of queries is large. In comparison, the complexity of WeakNAS scales almost linearly, as can be verified in Appendix Table~\ref{table:runtime}. In our experiments, we observe WeakNAS to perform empirically more competitively than current BO-based NAS methods at larger query numbers, besides being way more efficient. 

To further convince that WeakNAS is indeed an effective simplification compared to the explicit posterior modeling in BO, we report an apple-to-apple comparison, by use the same weak predictor from WeakNAS, plus obtaining its uncertainty estimation by calculating its variance using a deep ensemble of five model~\cite{lakshminarayanan2016simple}; we then use the classic Expected Improvement (EI)~\cite{jones1998efficient} acquisition function. Table~\ref{table:bo-variant} confirms that such BO variant of WeakNAS is inferior our proposed formulation.

\noindent\textbf{Advanced Architecture Encoding \citep{yan2020does, yan2021cate}} We also compare WeakNAS with NAS using custom architecture representation either in a unsupervised way such as arch2vec \citep{yan2020does}, or a supervised way such as CATE \cite{yan2021cate}. We show our WeakNAS could achieve comparable performance to both methods. Further, those architecture embedding are essentially \textbf{complementary} to our method to further boost the performance of WeakNAS, as shown in Appendix Section~\ref{sec:arch-encode}.

\noindent\textbf{LaNAS \citep{wang2019sampleefficient}:} LaNAS and our framework both follow the divide-and-conquer idea, yet with two methodological differences: \text{\textit{(a) How to split the search space}}:
LaNAS learns a \textit{classifier} to do binary ``hard'' partition on the search space (no ranking information utilized) and split it into two equally-sized subspaces. Ours uses a \textit{regressor} to regress the performance of sampled architectures, and utilizes the ranking information to sample a percentage of the top samples (``soft'' partition), with the sample size $N$ being controllable.
\text{\textit{(b) How to do exploration}}: LaNAS uses Upper Confidence Bound (UCB) to explore the search space by not always choosing the best subspace (left-most node) for sampling,
while ours always chooses the best subspace and explore new architectures by adaptive sampling within it, via adjusting the ratio $\epsilon=M/N$ to randomly sample $M$ models from Top $N$. Tables \ref{table:sota-nasbench101} and \ref{table:optimal} shows the superior sample efficiency of WeakNAS over LaNAS on NAS-Bench-101/201.

\noindent\textbf{Semi-NAS \citep{luo2020semi} and Semi-Assessor\cite{tang2020semi}:} Both our method and Semi-NAS/Semi-Assessor use an iterative algorithm containing prediction and sampling. The main difference lies in the use of pseudo labels:  Semi-NAS and Semi-Assessor use pseudo labels as noisy labels to augment the training set, therefore being able to leverage ``unlabeled samples" (e.g., architectures without true accuracies, but with only accuracies generated by the predictors) to update their predictors. Our method explores an \textbf{orthogonal} innovative direction, where the ``pseudo labels" generated by the current predictor guide our sampling procedure, but are \textbf{never used} for training the next predictor.

That said, we point out that our method can be \textbf{complementary} to those semi-supervised methods \citep{luo2020semi,tang2020semi}, thus they can further be \textbf{integrated} as one, For example, Semi-NAS can be used as a meta sampling method, where at each iteration we further train a Semi-NAS predictor with pseudo labeling strategy to augment the training set of our weak predictors. We show in Appendix Table~\ref{table:meta-sampling} that the combination of our method with Semi-NAS can further boost the performance of WeakNAS. 

\noindent\textbf{BRP-NAS \citep{chau2020brp}:} BRP-NAS uses a stronger GCN-based binary relation predictor which utilize extra topological prior, and leveraged a different scheme to control exploitation and exploration trade-off compare to our WeakNAS. Further, BRP-NAS also use a somehow unique setting, i.e. evaluating Top-40 predictions by the NAS predictor instead of the more common setting of Top-1 \citep{luo2018neural,shi2019bridging,wang2019sampleefficient,luo2020semi}. Therefore, we include our comparison to BRP-NAS and more details in Appendix Section~\ref{sec:compare-brp}.

\begin{table*}[!htp]
\centering
\scalebox{0.82}{
\begin{tabular}{lrccccr}
\toprule
Model & Queries(\#) & Top-1 Err.(\%) & Top-5 Err.(\%) & Params(M) & FLOPs(M) & GPU Days\\
\midrule
MobileNetV2  & -  & 25.3 &  -  & 6.9 & 585 & -\\
ShuffletNetV2 & - & 25.1 &  -  & 5.0 & 591 & -\\
\midrule
SNAS\citep{xie2018snas} & - & 27.3 & 9.2 & 4.3 & 522 & 1.5\\
DARTS\citep{liu2018darts}  & - & 26.9 & 9.0 & 4.9 & 595 & 4.0 \\
P-DARTS\citep{chen2019progressive}  & - & 24.4 & 7.4 & 4.9 & 557 & 0.3 \\
PC-DARTS\citep{xu2019pc}  & - & 24.2 & 7.3 & 5.3 & 597 & 3.8 \\
DS-NAS\citep{xu2019pc}  & - & 24.2 & 7.3 & 5.3 & 597 & 10.4 \\
\midrule
NASNet-A \citep{zoph2018learning} & 20000 & 26.0 & 8.4 & 5.3 & 564 & 2000\\
AmoebaNet-A \citep{real2019regularized}  & 10000 & 25.5 & 8.0 & 5.1 & 555 & 3150
\\
PNAS \citep{liu2018progressive}  & 1160 & 25.8 & 8.1 & 5.1 & 588 & 200 \\
NAO \citep{luo2018neural}  & 1000 & 24.5 & 7.8 & 6.5 & 590 & 200\\
\midrule
LaNAS \citep{wang2019sampleefficient} (Oneshot) & 800 & 24.1 & - & 5.4 & 567 & 3\\
LaNAS \citep{wang2019sampleefficient} & 800 & 23.5 & - & 5.1 & 570 & 150 \\
\midrule
\textbf{WeakNAS} & \textbf{800} & \textbf{23.5} & \textbf{6.8} & \textbf{5.5} & \textbf{591} & \textbf{2.5} \\
\bottomrule
\end{tabular}
}
\caption{Comparison to SOTA results on ImageNet using NASNet search space.}
\label{table:imagenet-nasnet}
\vspace{-1.2em}
\end{table*}

\begin{table}[!htp]
\centering
\scalebox{0.9}{
\begin{tabular}{lrcccr}
\toprule
Model & Queries(\#) & Top-1 Acc.(\%) & Top-5 Acc.(\%) & FLOPs(M) & GPU Days\textsuperscript{$\star$} \\ %
\midrule
Proxyless NAS\citep{cai2018proxylessnas} & - & 75.1 & 92.9 & - & -\\
Semi-NAS\citep{luo2020semi} & 300 & 76.5 & 93.2 & 599 & -\\
BigNAS\citep{yu2020bignas} & - & 76.5 & - & 586 & -\\
FBNetv3\citep{dai2020fbnetv3} & 20000 & 80.5 & 95.1 & 557 & -\\
OFA\citep{cai2019once} & 16000 & 80.0 & - & 595 & 1.6 \\
LaNAS\citep{wang2019sampleefficient} & 800 & 80.8 & - & 598 & 0.3 \\
\midrule
\multirow{2}{*}{\textbf{WeakNAS}} & \textbf{1000} & \textbf{81.3} & \textbf{95.1} & \textbf{560} & \textbf{0.16} \\
& \textbf{800} & \textbf{81.2} & \textbf{95.2} & \textbf{593} & \textbf{0.13} \\
\bottomrule
\end{tabular}}
\vspace{0.5em}
\caption{Comparison to SOTA results on ImageNet using MobileNet search space. \textsuperscript{$\star$}Does not include supernet training cost.}
\label{table:imagenet-mobilenet}
\vspace{-1em}
\end{table}

\vspace{-1em}
\section{Conclusions and Discussions of Broad Impact}
\label{sec:conclusion_boarder_impact}
\vspace{-0.5em}
In this paper, we present a novel predictor-based NAS framework named WeakNAS that progressively shrinks the sampling space, by learning a series of weak predictors that can connect towards the best architectures. By co-evolving the sampling stage and learning stage, our weak predictors can progressively evolve to sample towards the subspace of best architectures, thus greatly simplifying the learning task of each predictor. Extensive experiments on popular NAS benchmarks prove that the proposed method is both sample-efficient and robust to various combinations of predictors and architecture encoding means. However,  WeakNAS is still limited by the human-designed encoding of neural architectures, and our future work plans to investigate how to jointly learn the predictor and encoding in our framework.

For broader impact, the excellent sample-efficiency of WeakNAS reduces the resource and energy consumption needed to search for efficient models, while still maintaining SoTA performance. That can effectively serve the goal of GreenAI, from model search to model deployment. It might meanwhile be subject to the potential abuse of searching for models serving malicious purposes.

\vspace{-0.5em}
\section*{Acknowledgment}
\vspace{-0.5em}
Z.W. is in part supported by an NSF CCRI project (\#2016727).

\bibliographystyle{unsrt}
\bibliography{egbib}

\begin{thebibliography}{10}

\bibitem{liu2018darts}
Hanxiao Liu, Karen Simonyan, and Yiming Yang.
\newblock Darts: Differentiable architecture search.
\newblock {\em arXiv preprint arXiv:1806.09055}, 2018.

\bibitem{luo2018neural}
Renqian Luo, Fei Tian, Tao Qin, Enhong Chen, and Tie-Yan Liu.
\newblock Neural architecture optimization.
\newblock In {\em Advances in neural information processing systems}, pages
  7816--7827, 2018.

\bibitem{wu2019fbnet}
Bichen Wu, Xiaoliang Dai, Peizhao Zhang, Yanghan Wang, Fei Sun, Yiming Wu,
  Yuandong Tian, Peter Vajda, Yangqing Jia, and Kurt Keutzer.
\newblock Fbnet: Hardware-aware efficient convnet design via differentiable
  neural architecture search.
\newblock In {\em Proceedings of the IEEE Conference on Computer Vision and
  Pattern Recognition}, pages 10734--10742, 2019.

\bibitem{howard2019searching}
Andrew Howard, Mark Sandler, Grace Chu, Liang-Chieh Chen, Bo~Chen, Mingxing
  Tan, Weijun Wang, Yukun Zhu, Ruoming Pang, Vijay Vasudevan, et~al.
\newblock Searching for mobilenetv3.
\newblock In {\em Proceedings of the IEEE International Conference on Computer
  Vision}, 2019.

\bibitem{ning2020generic}
Xuefei Ning, Yin Zheng, Tianchen Zhao, Yu~Wang, and Huazhong Yang.
\newblock A generic graph-based neural architecture encoding scheme for
  predictor-based nas.
\newblock {\em arXiv preprint arXiv:2004.01899}, 2020.

\bibitem{wei2020npenas}
Chen Wei, Chuang Niu, Yiping Tang, and Jimin Liang.
\newblock Npenas: Neural predictor guided evolution for neural architecture
  search.
\newblock {\em arXiv preprint arXiv:2003.12857}, 2020.

\bibitem{wen2019neural}
Wei Wen, Hanxiao Liu, Hai Li, Yiran Chen, Gabriel Bender, and Pieter-Jan
  Kindermans.
\newblock Neural predictor for neural architecture search.
\newblock {\em arXiv preprint arXiv:1912.00848}, 2019.

\bibitem{chau2020brp}
Lukasz Dudziak, Thomas Chau, Mohamed Abdelfattah, Royson Lee, Hyeji Kim, and
  Nicholas Lane.
\newblock Brp-nas: Prediction-based nas using gcns.
\newblock {\em Advances in Neural Information Processing Systems}, 33, 2020.

\bibitem{luo2020neural}
Renqian Luo, Xu~Tan, Rui Wang, Tao Qin, Enhong Chen, and Tie-Yan Liu.
\newblock Neural architecture search with gbdt.
\newblock {\em arXiv preprint arXiv:2007.04785}, 2020.

\bibitem{wang2020dc}
Yunhe Wang, Yixing Xu, and Dacheng Tao.
\newblock Dc-nas: Divide-and-conquer neural architecture search.
\newblock {\em arXiv preprint arXiv:2005.14456}, 2020.

\bibitem{dai2020data}
Xiyang Dai, Dongdong Chen, Mengchen Liu, Yinpeng Chen, and Lu~Yuan.
\newblock Da-nas: Data adapted pruning for efficient neural architecture
  search.
\newblock {\em ECCV 2020}, 2020.

\bibitem{yang2020hournas}
Zhaohui Yang, Yunhe Wang, Xinghao Chen, Jianyuan Guo, Wei Zhang, Chao Xu,
  Chunjing Xu, Dacheng Tao, and Chang Xu.
\newblock Hournas: Extremely fast neural architecture search through an
  hourglass lens.
\newblock {\em arXiv preprint arXiv:2005.14446}, 2020.

\bibitem{zoph2016neural}
Barret Zoph and Quoc~V Le.
\newblock Neural architecture search with reinforcement learning.
\newblock {\em arXiv preprint arXiv:1611.01578}, 2016.

\bibitem{real2019regularized}
Esteban Real, Alok Aggarwal, Yanping Huang, and Quoc~V Le.
\newblock Regularized evolution for image classifier architecture search.
\newblock In {\em Proceedings of the aaai conference on artificial
  intelligence}, volume~33, pages 4780--4789, 2019.

\bibitem{yang2020cars}
Zhaohui Yang, Yunhe Wang, Xinghao Chen, Boxin Shi, Chao Xu, Chunjing Xu,
  Qi~Tian, and Chang Xu.
\newblock Cars: Continuous evolution for efficient neural architecture search.
\newblock In {\em Proceedings of the IEEE/CVF Conference on Computer Vision and
  Pattern Recognition}, pages 1829--1838, 2020.

\bibitem{hong2020dropnas}
Weijun Hong, Guilin Li, Weinan Zhang, Ruiming Tang, Yunhe Wang, Zhenguo Li, and
  Yong Yu.
\newblock Dropnas: Grouped operation dropout for differentiable architecture
  search.
\newblock In {\em International Joint Conference on Artificial Intelligence},
  2020.

\bibitem{xu2019renas}
Yixing Xu, Yunhe Wang, Kai Han, Shangling Jui, Chunjing Xu, Qi~Tian, and Chang
  Xu.
\newblock Renas: Relativistic evaluation of neural architecture search.
\newblock {\em arXiv preprint arXiv:1910.01523}, 2019.

\bibitem{li2020neural}
Yanxi Li, Minjing Dong, Yunhe Wang, and Chang Xu.
\newblock Neural architecture search in a proxy validation loss landscape.
\newblock In {\em International Conference on Machine Learning}, pages
  5853--5862. PMLR, 2020.

\bibitem{shi2019bridging}
Han Shi, Renjie Pi, Hang Xu, Zhenguo Li, James Kwok, and Tong Zhang.
\newblock Bridging the gap between sample-based and one-shot neural
  architecture search with bonas.
\newblock {\em Advances in Neural Information Processing Systems}, 33, 2020.

\bibitem{luo2020semi}
Renqian Luo, Xu~Tan, Rui Wang, Tao Qin, Enhong Chen, and Tie-Yan Liu.
\newblock Semi-supervised neural architecture search.
\newblock {\em arXiv preprint arXiv:2002.10389}, 2020.

\bibitem{wang2019sampleefficient}
Linnan Wang, Saining Xie, Teng Li, Rodrigo Fonseca, and Yuandong Tian.
\newblock Sample-efficient neural architecture search by learning actions for
  monte carlo tree search.
\newblock {\em IEEE Transactions on Pattern Analysis and Machine Intelligence},
  2021.

\bibitem{tappenden2016inexact}
Rachael Tappenden, Peter Richt{\'a}rik, and Jacek Gondzio.
\newblock Inexact coordinate descent: complexity and preconditioning.
\newblock {\em Journal of Optimization Theory and Applications},
  170(1):144--176, 2016.

\bibitem{schmidt2011convergence}
Mark Schmidt, Nicolas~Le Roux, and Francis Bach.
\newblock Convergence rates of inexact proximal-gradient methods for convex
  optimization.
\newblock {\em arXiv preprint arXiv:1109.2415}, 2011.

\bibitem{hager2020convergence}
William~W Hager and Hongchao Zhang.
\newblock Convergence rates for an inexact admm applied to separable convex
  optimization.
\newblock {\em Computational Optimization and Applications}, 2020.

\bibitem{zhou2012ensemble}
Zhi-Hua Zhou.
\newblock {\em Ensemble methods: foundations and algorithms}.
\newblock CRC press, 2012.

\bibitem{radosavovic2020designing}
Ilija Radosavovic, Raj~Prateek Kosaraju, Ross Girshick, Kaiming He, and Piotr
  Doll{\'a}r.
\newblock Designing network design spaces.
\newblock In {\em Proceedings of the IEEE/CVF Conference on Computer Vision and
  Pattern Recognition}, pages 10428--10436, 2020.

\bibitem{sutton2018reinforcement}
Richard~S Sutton and Andrew~G Barto.
\newblock {\em Reinforcement learning: An introduction}.
\newblock MIT press, 2018.

\bibitem{siems2020bench}
Julien Siems, Lucas Zimmer, Arber Zela, Jovita Lukasik, Margret Keuper, and
  Frank Hutter.
\newblock Nas-bench-301 and the case for surrogate benchmarks for neural
  architecture search.
\newblock {\em arXiv preprint arXiv:2008.09777}, 2020.

\bibitem{chen2016xgboost}
Tianqi Chen and Carlos Guestrin.
\newblock Xgboost: A scalable tree boosting system.
\newblock In {\em Proceedings of the 22nd acm sigkdd international conference
  on knowledge discovery and data mining}, pages 785--794, 2016.

\bibitem{hu2020angle}
Yiming Hu, Yuding Liang, Zichao Guo, Ruosi Wan, Xiangyu Zhang, Yichen Wei,
  Qingyi Gu, and Jian Sun.
\newblock Angle-based search space shrinking for neural architecture search.
\newblock {\em arXiv preprint arXiv:2004.13431}, 2020.

\bibitem{Dong2020NAS-Bench-201:}
Xuanyi Dong and Yi~Yang.
\newblock Nas-bench-201: Extending the scope of reproducible neural
  architecture search.
\newblock In {\em International Conference on Learning Representations}, 2020.

\bibitem{ying2019bench}
Chris Ying, Aaron Klein, Eric Christiansen, Esteban Real, Kevin Murphy, and
  Frank Hutter.
\newblock Nas-bench-101: Towards reproducible neural architecture search.
\newblock In {\em International Conference on Machine Learning}, pages
  7105--7114, 2019.

\bibitem{krizhevsky2009learning}
Alex Krizhevsky, Geoffrey Hinton, et~al.
\newblock Learning multiple layers of features from tiny images.
\newblock 2009.

\bibitem{chrabaszcz2017downsampled}
Patryk Chrabaszcz, Ilya Loshchilov, and Frank Hutter.
\newblock A downsampled variant of imagenet as an alternative to the cifar
  datasets.
\newblock {\em arXiv preprint arXiv:1707.08819}, 2017.

\bibitem{zoph2018learning}
Barret Zoph, Vijay Vasudevan, Jonathon Shlens, and Quoc~V Le.
\newblock Learning transferable architectures for scalable image recognition.
\newblock In {\em Proceedings of the IEEE conference on computer vision and
  pattern recognition}, pages 8697--8710, 2018.

\bibitem{cai2019once}
Han Cai, Chuang Gan, Tianzhe Wang, Zhekai Zhang, and Song Han.
\newblock Once-for-all: Train one network and specialize it for efficient
  deployment.
\newblock {\em arXiv preprint arXiv:1908.09791}, 2019.

\bibitem{deng2009imagenet}
Jia Deng, Wei Dong, Richard Socher, Li-Jia Li, Kai Li, and Li~Fei-Fei.
\newblock Imagenet: A large-scale hierarchical image database.
\newblock In {\em 2009 IEEE conference on computer vision and pattern
  recognition}, pages 248--255. Ieee, 2009.

\bibitem{guo2019single}
Zichao Guo, Xiangyu Zhang, Haoyuan Mu, Wen Heng, Zechun Liu, Yichen Wei, and
  Jian Sun.
\newblock Single path one-shot neural architecture search with uniform
  sampling.
\newblock {\em arXiv preprint arXiv:1904.00420}, 2019.

\bibitem{rw2019timm}
Ross Wightman.
\newblock Pytorch image models.
\newblock \url{https://github.com/rwightman/pytorch-image-models}, 2019.

\bibitem{wang2019alphax}
Linnan Wang, Yiyang Zhao, Yuu Jinnai, Yuandong Tian, and Rodrigo Fonseca.
\newblock Alphax: exploring neural architectures with deep neural networks and
  monte carlo tree search.
\newblock {\em arXiv preprint arXiv:1903.11059}, 2019.

\bibitem{yan2020does}
Shen Yan, Yu~Zheng, Wei Ao, Xiao Zeng, and Mi~Zhang.
\newblock Does unsupervised architecture representation learning help neural
  architecture search?
\newblock {\em Advances in Neural Information Processing Systems}, 33, 2020.

\bibitem{tang2020semi}
Yehui Tang, Yunhe Wang, Yixing Xu, Hanting Chen, Boxin Shi, Chao Xu, Chunjing
  Xu, Qi~Tian, and Chang Xu.
\newblock A semi-supervised assessor of neural architectures.
\newblock In {\em Proceedings of the IEEE/CVF Conference on Computer Vision and
  Pattern Recognition}, pages 1810--1819, 2020.

\bibitem{yan2021cate}
Shen Yan, Kaiqiang Song, Fei Liu, and Mi~Zhang.
\newblock Cate: Computation-aware neural architecture encoding with
  transformers.
\newblock {\em arXiv preprint arXiv:2102.07108}, 2021.

\bibitem{white2019bananas}
Colin White, Willie Neiswanger, and Yash Savani.
\newblock Bananas: Bayesian optimization with neural architectures for neural
  architecture search.
\newblock In {\em Proceedings of the AAAI Conference on Artificial
  Intelligence}, volume~35, pages 10293--10301, 2021.

\bibitem{ru2020interpretable}
Binxin Ru, Xingchen Wan, Xiaowen Dong, and Michael Osborne.
\newblock Interpretable neural architecture search via bayesian optimisation
  with weisfeiler-lehman kernels.
\newblock In {\em International Conference on Learning Representations}, 2021.

\bibitem{liu2018progressive}
Chenxi Liu, Barret Zoph, Maxim Neumann, Jonathon Shlens, Wei Hua, Li-Jia Li,
  Li~Fei-Fei, Alan Yuille, Jonathan Huang, and Kevin Murphy.
\newblock Progressive neural architecture search.
\newblock In {\em Proceedings of the European Conference on Computer Vision
  (ECCV)}, pages 19--34, 2018.

\bibitem{yu2020bignas}
Jiahui Yu, Pengchong Jin, Hanxiao Liu, Gabriel Bender, Pieter-Jan Kindermans,
  Mingxing Tan, Thomas Huang, Xiaodan Song, Ruoming Pang, and Quoc Le.
\newblock Bignas: Scaling up neural architecture search with big single-stage
  models.
\newblock In {\em European Conference on Computer Vision}, pages 702--717.
  Springer, 2020.

\bibitem{dai2020fbnetv3}
Xiaoliang Dai, Alvin Wan, Peizhao Zhang, Bichen Wu, Zijian He, Zhen Wei, Kan
  Chen, Yuandong Tian, Matthew Yu, Peter Vajda, et~al.
\newblock Fbnetv3: Joint architecture-recipe search using neural acquisition
  function.
\newblock {\em arXiv preprint arXiv:2006.02049}, 2020.

\bibitem{lakshminarayanan2016simple}
Balaji Lakshminarayanan, Alexander Pritzel, and Charles Blundell.
\newblock Simple and scalable predictive uncertainty estimation using deep
  ensembles.
\newblock {\em arXiv preprint arXiv:1612.01474}, 2016.

\bibitem{jones1998efficient}
Donald~R Jones, Matthias Schonlau, and William~J Welch.
\newblock Efficient global optimization of expensive black-box functions.
\newblock {\em Journal of Global optimization}, 13(4):455--492, 1998.

\bibitem{xie2018snas}
Sirui Xie, Hehui Zheng, Chunxiao Liu, and Liang Lin.
\newblock Snas: stochastic neural architecture search.
\newblock {\em arXiv preprint arXiv:1812.09926}, 2018.

\bibitem{chen2019progressive}
Xin Chen, Lingxi Xie, Jun Wu, and Qi~Tian.
\newblock Progressive differentiable architecture search: Bridging the depth
  gap between search and evaluation.
\newblock In {\em Proceedings of the IEEE International Conference on Computer
  Vision}, pages 1294--1303, 2019.

\bibitem{xu2019pc}
Yuhui Xu, Lingxi Xie, Xiaopeng Zhang, Xin Chen, Guo-Jun Qi, Qi~Tian, and
  Hongkai Xiong.
\newblock Pc-darts: Partial channel connections for memory-efficient
  differentiable architecture search.
\newblock {\em arXiv preprint arXiv:1907.05737}, 2019.

\bibitem{cai2018proxylessnas}
Han Cai, Ligeng Zhu, and Song Han.
\newblock Proxylessnas: Direct neural architecture search on target task and
  hardware.
\newblock {\em arXiv preprint arXiv:1812.00332}, 2018.

\bibitem{luo2020accuracy}
Renqian Luo, Xu~Tan, Rui Wang, Tao Qin, Enhong Chen, and Tie-Yan Liu.
\newblock Accuracy prediction with non-neural model for neural architecture
  search.
\newblock {\em arXiv preprint arXiv:2007.04785}, 2020.

\end{thebibliography}

\clearpage

\appendix

\section{Implementation details of baselines methods}

For random search and regularized evolution\cite{real2019regularized} baseline, we use the public implementation from this link\footnote{\url{https://github.com/D-X-Y/AutoDL-Projects}}. For random search, we selection 100 random architectures at each iteration. For regularized evolution, We set the initial population to 10, and the sample size each iteration to 3.

\section{Runtime comparsion of WeakNAS}

We show the runtime comparison of WeakNAS and its BO variant in Table~\ref{table:runtime}. We can see the BO variant is much slower in training proxy models due the ensembling of multiple models. Moreover, it's also several magnitude slower when deriving new samples, due to the calculation of its Expected Improvement (EI) acquisition function \cite{jones1998efficient} being extremely costly.

\begin{table*}[!htp]
\centering
\scalebox{0.7}{\begin{tabular}{llcrccr}
\toprule
 Method & Predictors & Config & Train proxy model (s/arch) & Derive new samples (s/arch)  \\
\midrule
\multirow{3}{*}{WeakNAS} & MLP & 4 layers @1000 hidden & $8.59\times10^{-5}$ & $3.53\times10^{-5}$ \\
 & Gradient Boosting Tree & 1000 Trees & $5.70\times10^{-4}$ & $5.54\times10^{-7}$ \\
 & Random Forrest & 1000 Forests & $3.20\times10^{-3}$ & $1.77\times10^{-4}$ \\
\midrule
WeakNAS (BO Variant) & 5 x MLPs & EI acquisition & $2.84\times10^{-4}$ & $1.32\times10^{-1}$ \\
\bottomrule
\end{tabular}}
\caption{Runtime Comparsion of WeakNAS}
\label{table:runtime}
\end{table*}

\section{Ablation on the architecture encoding}
\label{sec:arch-encode}

We compare the effect of using different architecture encodings in in Table~\ref{table:cate}. We found when combined with CATE embedding \cite{yan2021cate}, the performance of WeakNAS can be further improved, compared to WeakNAS baseline with adjacency matrix encoding used in \cite{ying2019bench}. This also leads to stronger performance than cate-DNGO-LS baseline in CATE~\cite{yan2021cate}, which demonstrates that CATE embedding~\cite{yan2021cate} is an orthogonal contribution to WeakNAS, and they are mutually compatible.
\begin{table*}[!htp]
\centering
\scalebox{0.89}{\begin{tabular}{lcrccr}
\toprule
 Methods &\#Queries &  Test Acc.(\%) & SD(\%) & Test Regret(\%) & Avg. Rank\\
 \midrule
 CATE (cate-DNGO-LS)\citep{yan2021cate} & 150 & 94.10 & - & 0.22 & 12.3 \\
WeakNAS + Adjacency matrix\cite{ying2019bench} & 150 & 94.10 & 0.19 & 0.22 & 12.3 \\
WeakNAS + CATE\cite{yan2021cate} & 150 & 94.19 & 0.12 & 0.13 & 5.24 \\
\bottomrule
\end{tabular}}
\caption{Details on Ablation on meta-sampling methods on NAS-Bench-101}
\label{table:cate}
\vspace{-1em}
\end{table*}

\section{Ablation on number of initial samples}
\label{sec:num_init}
We conduct a controlled experiment in varying the number of initial samples $|M_{0}|$ in Table~\ref{table:init}. On NAS-Bench-101, we vary $|M_{0}|$ from 10 to 200, and found a "warm start" with good initial samples is crucial for good performance. Too small number of $|M_{0}|$ might makes the predictor lose track of the good performing regions. As shown in Table~\ref{table:init}. We empirically found $|M_{0}|$=100 can ensure highly stable performance on NAS-Bench-101.

\begin{table*}[!htp]
\centering

\scalebox{0.90}{\begin{tabular}{lrrrrr}
\toprule
$|M_{0}|$ &\#Queries &  Test Acc.(\%) & SD(\%) & Test Regret(\%) & Avg. Rank\\
\midrule
10 & 1000 & 94.14 & 0.10 & 0.18 & 9.1 \\
\textbf{100} & \textbf{1000} & \textbf{94.25} & \textbf{0.04} & \textbf{0.07} & \textbf{1.7}\\
200 & 1000 & 94.19 & 0.08 & 0.13 & 5.2 \\
\midrule
10 & 200 & 94.04 & 0.13 & 0.28 & 33.5 \\
\textbf{100} & \textbf{200} & \textbf{94.18} & \textbf{0.14} & \textbf{0.14} & \textbf{5.6} \\
200 & 200 & 93.78 & 1.45 & 0.54 & 558.0 \\
\midrule
Optimal & - & 94.32 & - & 0.00 & 1.0 \\
\bottomrule
\end{tabular}}
\caption{Ablation on number of initial samples ${M}_{0}$ on NAS-Bench-101}
\label{table:init}
\vspace{-1em}
\end{table*}

\section{More comparison on NAS-Bench-201}

\begin{figure*}[!htp]
\begin{minipage}[]{1.0\textwidth}
\centering
\subfloat[CIFAR10]{\includegraphics[width=0.34\textwidth]{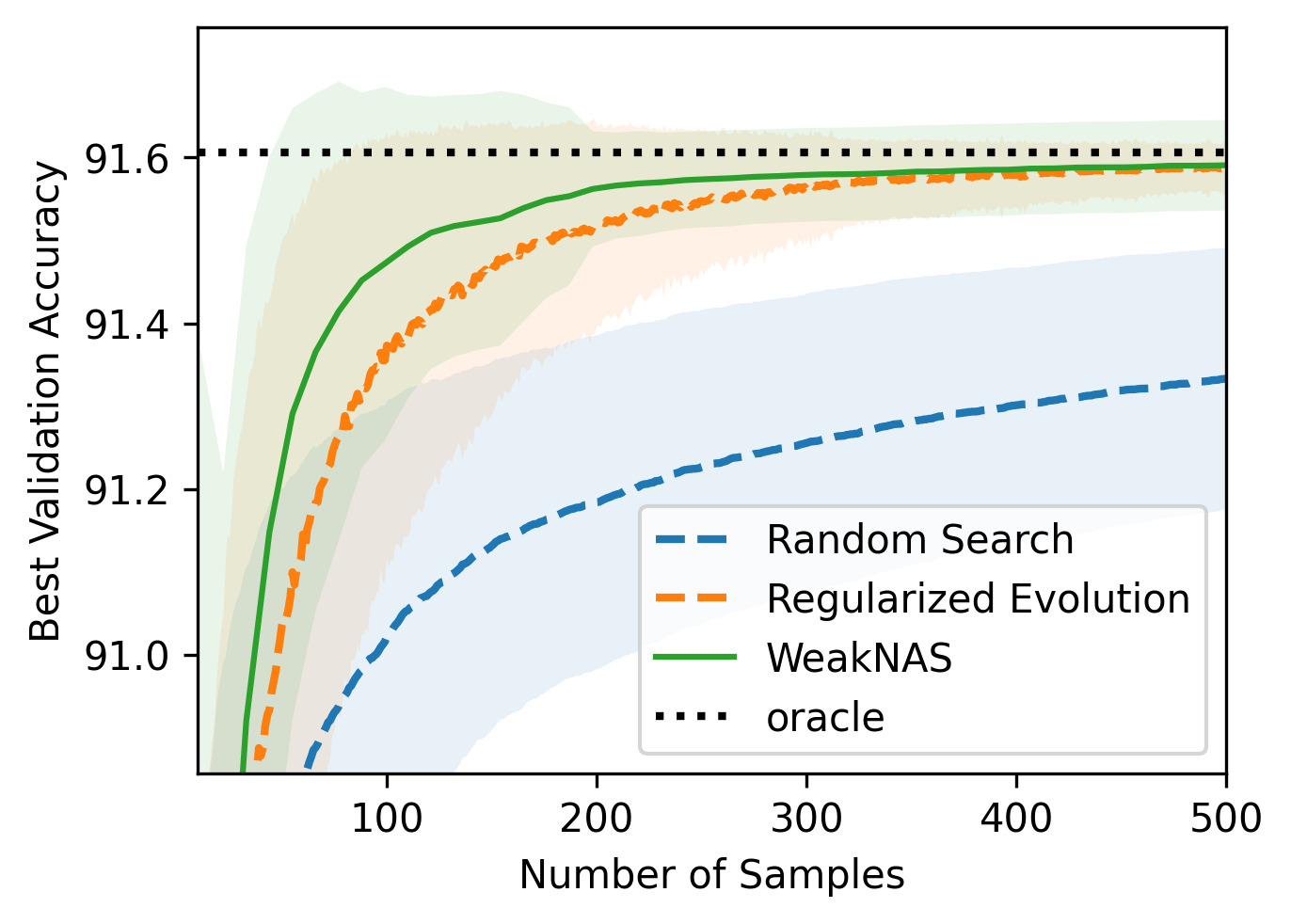}}
\subfloat[CIFAR100]{\includegraphics[width=0.34\textwidth]{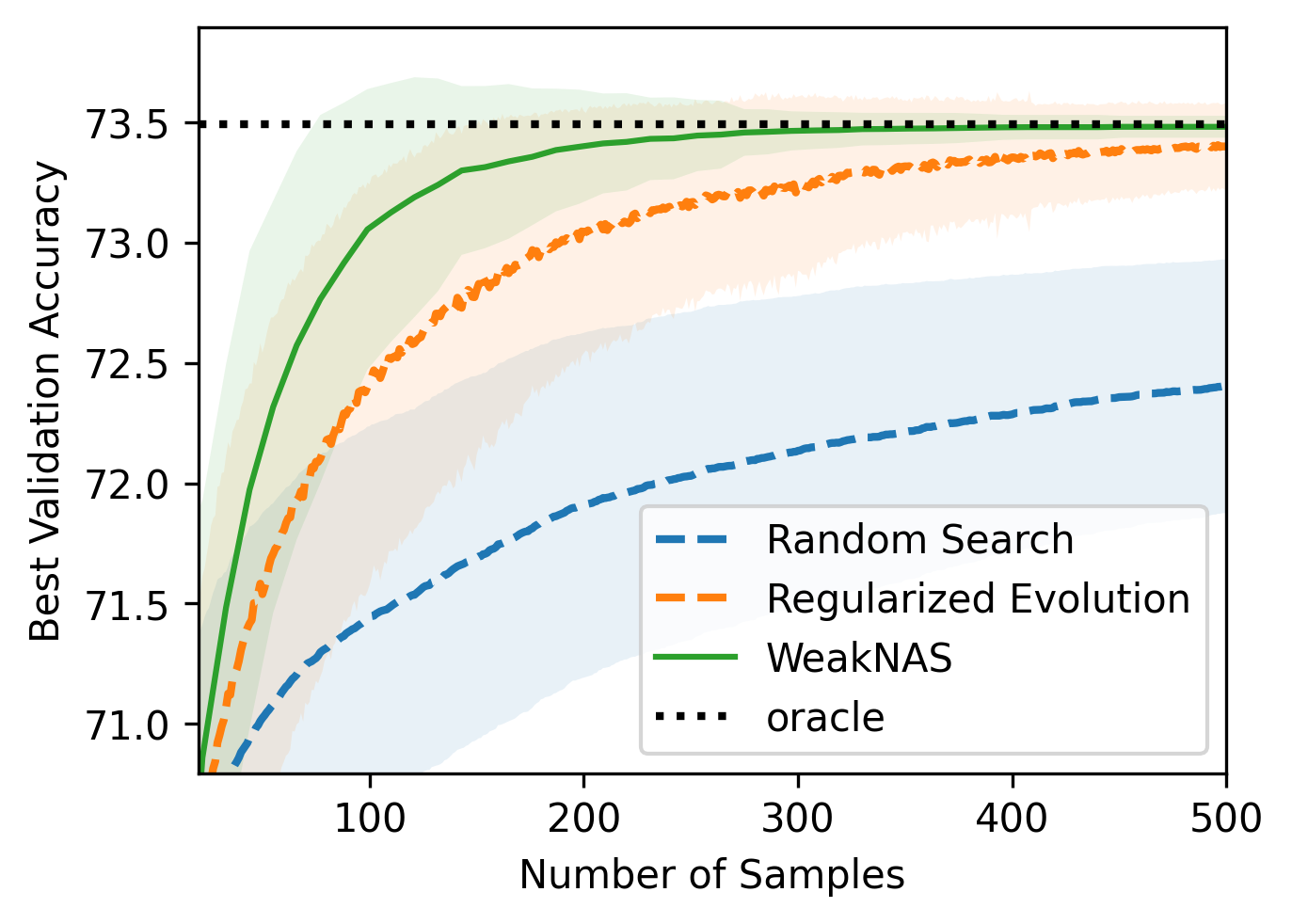}}
\subfloat[ImageNet16-120]{\includegraphics[width=0.34\textwidth]{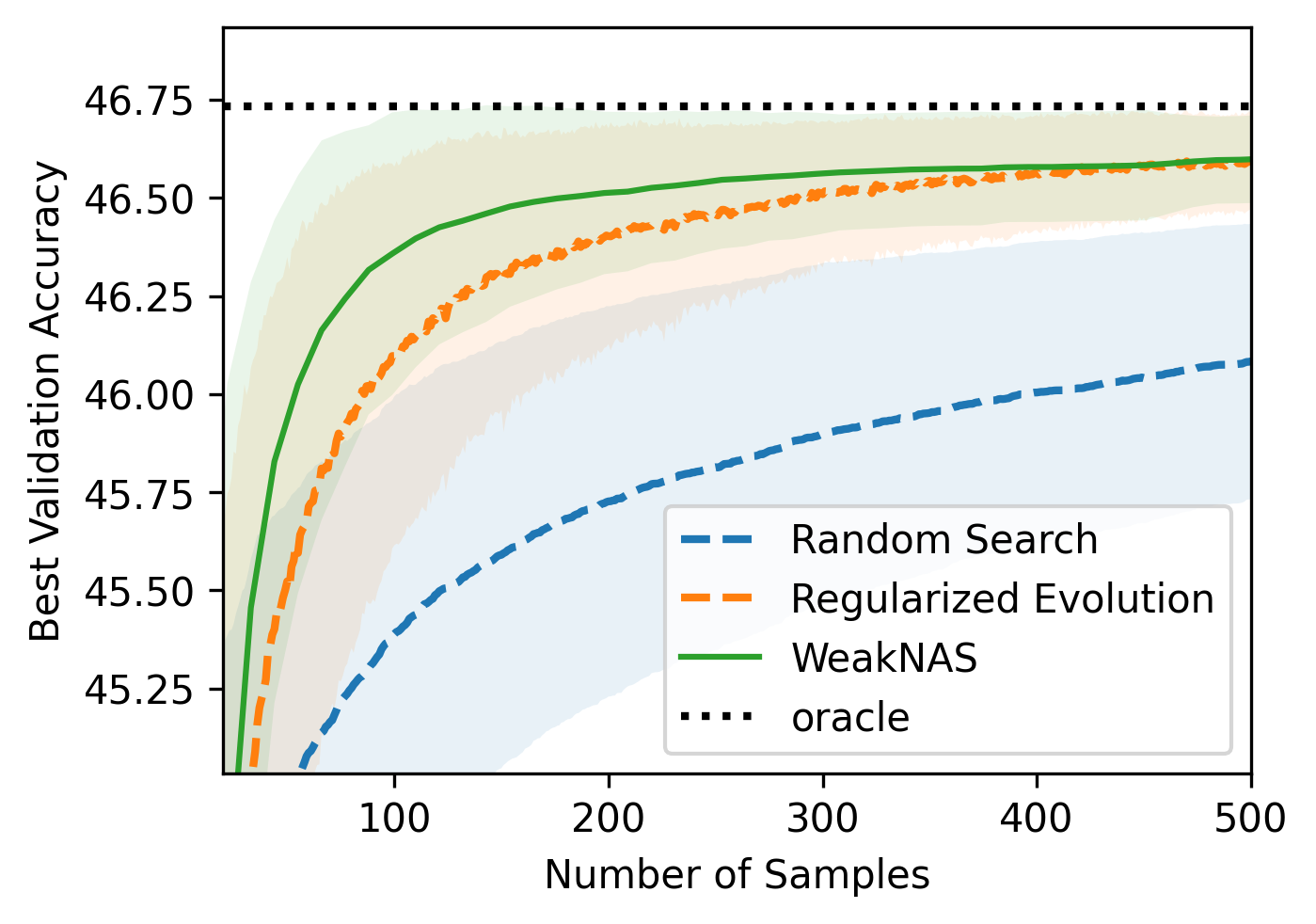}}\\
\vspace{-0.5em}
\caption{Comparison to SOTA on NAS-Bench-201 by varying number of samples. Solid lines and shadow regions denote the mean and std, respectively.}
\vspace{-0.5em}

\label{fig:sota-nasbench201}
\end{minipage}
\vspace{-0.5em}
\end{figure*}

We conduct a controlled experiment on NAS-Bench-201 by varying number of samples. As shown in Figure \ref{fig:sota-nasbench201}, our average performance over different number of samples is clearly better than Regularized Evolution \citep{real2019regularized} in all three subsets, with better stability indicated by confidence intervals.

\section{Comparison to BRP-NAS}
\label{sec:compare-brp}

\noindent\textbf{Evaluation strategy}: BRP-NAS\citep{chau2020brp} uses a unique setting that differs from other predictor-based NAS, i.e., evaluating Top 40 predictions by the NAS predictor instead of Top 1 prediction, and the later was commonly followed by others\cite{luo2018neural,shi2019bridging,wang2019sampleefficient,luo2020semi} and WeakNAS.

\noindent\textbf{Sampling strategy}: WeakNAS uses a different sampling strategy than that of BRP-NAS, given a sample budget of $M$, BRP-NAS picks both samples from Top-$K$ and $(M-K)$ random models from the entire search space, while our WeakNAS only picks $M$ random models in Top-$N$, thus is a more “greedy” strategy. BRP-NAS controls the exploitation and exploration trade-off by adjusting $\alpha={(M-K)}/{M}$, however they did not have any ablation discussing the exploitation and exploration trade-off and only empirically choose $\alpha=0.5$ as the default ratio. Our WeakNAS instead controls the exploitation and exploration trade-off by adjusting ${N}/{M}$ ratio, and we did a comprehensive analysis on the exploitation and exploration trade-off on both NAS-Bench and MobileNet Search Space on ImageNet in Section~\ref{sec:exploitation-exploration}.

\noindent\textbf{NAS predictor}: BRP-NAS uses a stronger GCN-based binary relation predictors which utilizes extra topological prior, on the other hand, our framework generalizes to all choices of predictors, including MLP, Regression Tree and Random Forest, thus is not picky on the choice of predictors.

To fairly compare with BRP-NAS, we follow the exact same setting for our WeakNAS predictor, e.g., incorporating the same graph convolutional network (GCN) based predictor and using Top-40 evaluation. As shown in Table~\ref{table:brp}, at 100 training samples, WeakNAS can achieve comparable performance to BRP-NAS \citep{chau2020brp}.
\begin{table*}[!htp]
\centering
\scalebox{0.85}{
\begin{tabular}{lrrcccc}
\toprule
Method & \#Train  & \#Queries &  Test Acc.(\%) & SD(\%) & Test Regret(\%)  & Avg. Rank \\
\midrule
BRP-NAS \citep{chau2020brp}  & 100 & 140 & 94.22 & - & 0.10 & 3.0\\
\midrule
\textbf{WeakNAS} & \textbf{100} & \textbf{140} & \textbf{94.23} & \textbf{0.09} & \textbf{0.09} & \textbf{2.3}\\
\midrule
Optimal & -  & - & 94.32 & - & 0.00 & 1.0 \\
\bottomrule
\end{tabular}}
\caption{Comparison to BRP-NAS on NAS-Bench-101.}
\label{table:brp}
\end{table*}

\section{Comparsion of meta-sampling methods in WeakNAS}
\label{sec:meta-sampling}

We also show that local search algorithm (hill climbing) or Semi-NAS \citep{luo2020semi} can be used as a meta sampling method in WeakNAS, which could further boost the performance of WeakNAS, here are the implementation details.

\noindent\textbf{Local Search} Given a network architecture embedding $s$ in NAS-Bench-101 Search Space, we first define a nearest neighbour function $N(s)$ as architecture that differ from $s$ by a edge or a operation. At each iteration, we random sample a initial sample $s_{i}$ from TopN predictions $\text{Top}_N(\tilde{P}^{k})$ and sample all of its nearest neighbour architecture in $N(v_{0})$. We then let the new $s_{i+1} = \argmax_{s \in N(s_{i})} f(s)$. We repeat the process iteratively until we reach a local maximum such that $\forall v \in N(s), f(s) \geqslant f(v)$ or the sampling budget $M$ of the iteration is reached.

\noindent\textbf{Semi-NAS} At the sampling stage of each iteration in WeakNAS, we further use Semi-NAS as a meta-sampling methods. Given a meta search space of 1000 architectures and a sample budget of 100 queries each iteration. We following the setting in Semi-NAS, using the same 4-layer MLP NAS predictor in WeakNAS and uses pseudo labels as noisy labels to augment the training set, therefore we are able to leverage ``unlabeled samples" (e.g., architectures with accuracy generated by the predictors) to update the predictor. We set the initial sample to be 10, and sample 10 more samples each iteration. Note that at the start of ${k}$-th WeakNAS iteration, we inherent the weight of Semi-NAS predictor from the previous (${k}$-1)-th WeakNAS iteration.

\begin{table*}[!htp]
\centering
\scalebox{0.89}{\begin{tabular}{lcrccccr}
\toprule
 Sampling (M from TopN) & M & N &\#Queries &  Test Acc.(\%) & SD(\%) & Test Regret(\%) & Avg. Rank\\
\midrule
WeakNAS & 100 & 1000 & 1000 & 94.25 & 0.04 & 0.07 & 1.7 \\
\midrule
Local Search & - & - & 1000 & 94.24 & 0.03 & 0.08 & 1.9 \\
Semi-NAS & - & - & 1000 & 94.26 & 0.02 & 0.06 & 1.6 \\
\bottomrule
\end{tabular}}
\caption{Ablation on meta-sampling methods on NAS-Bench-101}
\label{table:meta-sampling}
\end{table*}

\section{Details of Implementation on Open Domain Search Space}
\label{opendomaindetails}

We extend WeakNAS to open domain settings by (a) Construct the evaluation pool $\bar{X}$ by uniform sampling the whole search space ${X}$ (b) Apply WeakNAS in the evaluation space $\bar{X}$ to find the best performer. (c) Train the best performer architecture from scratch.

For instance, when working with MobileNet search space that includes $\approx10^{18}$ architectures, we uniformly sample 10K models as an evaluation pool, and further apply WeakNAS with a sample budget of 800 or 1000. When working with NASNet search space that includes $\approx10^{21}$ architectures, we uniformly sample 100K models as an evaluation pool, and further apply WeakNAS with a sample budget of 800.

In the following part, we take MobileNet open domain search space as a example, however we follow a similar procedure for NASNet search space.

\noindent\textbf{(a) Construct the evaluation pool $\bar{X}$ from the search space ${X}$}
We uniformly sample an evaluation pool to handle the extremely large MobileNet search space (${|X|}\approx10^{18}$), since its not doable to predict the performance of all architectures in ${X}$. We use uniform sampling due to a recent study \cite{radosavovic2020designing} reveal that human-designed NAS search spaces usually contain a fair proportion of good models compared to random design spaces, for example, in Figure 9 of \cite{radosavovic2020designing}, it shows that in NASNet/Amoeba/PNAS/ENAS/DARTS search spaces, Top 5\% of models only have a <1\% performance gap to the global optima. In practice, the uniform sampling strategy has been widely verified as effective in other works of predictor-based NAS such as \cite{wen2019neural, luo2020accuracy, dai2020fbnetv3}, For example, \cite{wen2019neural} \cite{luo2020accuracy}\cite{dai2020fbnetv3} set 
 to be 112K, 15K, 20K in a search space of ${10}^{18}$ networks. In our case, we set $|\bar{X}|$ = 10K.
 
\noindent\textbf{(b) Apply WeakNAS in the evaluation space $\bar{X}$}
We then further apply WeakNAS in the evaluation pool $\bar{X}$. This is because even with the evaluation pool $|\bar{X}|$ = 10K, it still takes days to evaluate all those models on ImageNet (in a weight-sharing SuperNet setting). Since the evaluation pool $\bar{X}$ was uniformly sampled from NAS search space $X$, it preserves the highly-structured nature of $X$. As a result, we can leverage WeakNAS to navigate through the highly-structured search space. WeakNAS build a iterative process, where it searches for some top-performing cluster at the initial search iteration and then “zoom-in” the cluster to find the top performers within the same cluster (as shown in Figure~\ref{fig:tsne}). At $k-th$ iteration, WeakNAS balance the exploration and exploitation trade-off by sampling 100 models from the Top 1000 predictions of the predictor $\tilde{f}^{k}$, it use the promising samples to further improve performance of the predictor in the next iteration $\tilde{f}^{k+1}$. We leverage WeakNAS to further decrease the number of queries to find the optimal in $\bar{X}$ by 10 times, the search cost has dropped from 25 GPU hours (evaluate all 10K samples in random evaluation pool) to 2.5 GPU hours (use WeakNAS in 10K random evaluation pool), while still achieving a solid performance of 81.3\% on ImageNet (MobileNet Search Space).

\noindent\textbf{(c) Train the best performer architecture from scratch.} We follow a similar setting in LaNAS\cite{wang2019sampleefficient}, where we use Random Erase and RandAug, a drop out rate of 0.3 and a drop path rate of 0.0, we also use exponential moving average (EMA) with a decay rate of 0.9999. During training and evaluation, we set the image size to be 236x236 (In NASNet search space, we set the image size to be 224x224). We train for 300 epochs with warm-up of 3 epochs, we use a batch size of 1024 and RMSprop as the optimizer. We use a cosine decay learning rate scheduler with a starting learning rate of 1e-02 and a terminal learning rate of 1e-05.

\section{Ablation on exploitation-exploration trade-off in Mobilenet Search space on ImageNet}

For the ablation on open-domain search space, we follow the same setting in the Section~\ref{opendomaindetails}, however due to the prohibitive cost of training model from scratch in Section \ref{opendomaindetails} (c), we directly use accuracy derived from supernet.

WeakNAS uniformly samples M samples from TopN predictions at each iteration, thus we can adjust N/M ratio to balance the exploitation-exploration trade-off. In Table~\ref{table:exploration-mobilenet}, we set the total number of queries at 100, fix $M$ at 10 and while adjusting $N$ from 10 (more exploitation) to 1000 (more exploration), and use optimal in the 10K evaluation pool to measure the ranking and test regret. 
We found WeakNAS is quite robust within the range where N/M = 2.5 - 10, achieving the best performance at the sweet spot of N/M = 5. However, its performance drops significantly (by rank), while doing either too much exploitation (N/M <2.5) or too much exploration (N/M >25).

\begin{table*}[!htp]
\centering
\scalebox{0.86}{\begin{tabular}{lrrrrrrrr}
\toprule
 Sampling methods & M & TopN &\#Queries & SuperNet Test Acc.(\%) & SD(\%) & Test Regret(\%) & Avg. Rank\\
\midrule
Uniform	& - & - & 100 & 79.0609 & 0.0690 & 0.1671 & 94.58\\
\hline
\multirow{8}{*}{Iterative} & 10 & 10 & 100 & 79.1552 & 0.0553 & 0.0728 & 20.69\\
 & 10 & 25 & 100 & 79.1936 & 0.0289 & 0.0344 & 4.68\\
 & \textbf{10} & \textbf{50} & \textbf{100} & \textbf{79.2005} & \textbf{0.0300} & \textbf{0.0275} & \textbf{4.05}\\
 & 10 & 100 & 100 & 79.1954 & 0.0300 & 0.0326 & 4.63\\
 & 10 & 250 & 100 & 79.1755 & 0.0416 & 0.0525 & 10.58\\
 & 10 & 500 & 100 & 79.1710 & 0.0388 & 0.0570 & 10.80\\
 & 10 & 1000 & 100 & 79.1480 & 0.0459 & 0.0800 & 19.70\\
 & 10 & 2500 & 100 & 79.1274 & 0.0597 & 0.1006 & 33.64\\
\bottomrule
\end{tabular}}
\caption{Ablation on exploitation-exploration trade-off over 100 runs on MobleNet Search Space over ImageNet}
\label{table:exploration-mobilenet}
\end{table*}

\section{Founded Architecture on Open Domain Search}

We show the best architecture founded by WeakNAS with 800/1000 queries in Table \ref{table:imagenet-mobilenet-arch}.

\begin{table*}[!htp]
\scalebox{0.787}{
\begin{tabular}{c|c|c|c|c}
\toprule
\textbf{Id} & \textbf{Block} & \textbf{Kernel} & \textbf{\#Out Channel} & \textbf{Expand Ratio} \\
\hline
\multicolumn{5}{c}{WeakNAS @ 593 MFLOPs, \#Queries=800}\\
\hline
0 & Conv & 3 & 24 & - \\
1 & IRB & 3 & 24 & 1 \\
\hline
2 & IRB & 3 & 32 & 4 \\
3 & IRB & 5 & 32 & 6 \\
\hline
4 & IRB & 7 & 48 & 4 \\
5 & IRB & 5 & 48 & 3 \\
6 & IRB & 7 & 48 & 4 \\
7 & IRB & 3 & 48 & 6 \\
\hline
8  & IRB & 3 & 96 & 4 \\
9  & IRB & 7 & 96 & 6 \\
10 & IRB & 5 & 96 & 6 \\
11 & IRB & 7 & 96 & 3 \\
\hline
12 & IRB & 3 & 136 & 6 \\
13 & IRB & 3 & 136 & 6 \\
14 & IRB & 5 & 136 & 6 \\
15 & IRB & 5 & 136 & 3 \\
\hline
16 & IRB & 7 & 192 & 6 \\
17 & IRB & 5 & 192 & 6 \\
18 & IRB & 3 & 192 & 4 \\
19 & IRB & 5 & 192 & 3 \\
\hline
20 & Conv & 1 & 192 & - \\
21 & Conv & 1 & 1152 & - \\
22 & FC   & - & 1536 & - \\
\bottomrule
\end{tabular}
}
\scalebox{0.82}{
\begin{tabular}{c|c|c|c|c}
\toprule
\textbf{Id} & \textbf{Block} & \textbf{Kernel} & \textbf{\#Out Channel} & \textbf{Expand Ratio} \\
\hline
\multicolumn{5}{c}{WeakNAS @ 560 MFLOPs, \#Queries=1000}\\
\hline
0 & Conv & 3 & 24 & - \\
1 & IRB & 3 & 24 & 1 \\
\hline
2 & IRB & 5 & 32 & 3 \\
3 & IRB & 3 & 32 & 3 \\
4 & IRB & 3 & 32 & 4 \\
5 & IRB & 3 & 32 & 3 \\
\hline
6 & IRB & 5 & 48 & 4 \\
7 & IRB & 5 & 48 & 6 \\
8 & IRB & 5 & 48 & 4 \\
\hline
9  & IRB & 7 & 96 & 4 \\
10 & IRB & 5 & 96 & 6 \\
11 & IRB & 7 & 96 & 6 \\
\hline
12 & IRB & 3 & 136 & 6 \\
13 & IRB & 5 & 136 & 6 \\
14 & IRB & 5 & 136 & 6 \\
\hline
15 & IRB & 7 & 192 & 6 \\
16 & IRB & 5 & 192 & 6 \\
17 & IRB & 3 & 192 & 6 \\
18 & IRB & 5 & 192 & 3 \\
\hline
19 & Conv & 1 & 192 & - \\
20 & Conv & 1 & 1152 & - \\
21 & FC   & - & 1536 & - \\
\bottomrule
\end{tabular}
}
\caption{Neural architecture found by WeakNAS on ImageNet using MobileNet search space, i.e. results in main paper Table~6}
\label{table:imagenet-mobilenet-arch}
\end{table*}

\end{document}